\documentclass[a4paper,fleqn]{cas-dc}

\usepackage[numbers]{natbib}
\usepackage{gensymb}
\usepackage{threeparttable}
\usepackage{pifont}
\usepackage{color, todonotes}

\def\tsc#1{\csdef{#1}{\textsc{\lowercase{#1}}\xspace}}
\tsc{WGM}
\tsc{QE}
\tsc{EP}
\tsc{PMS}
\tsc{BEC}
\tsc{DE}

\begin{document}
\let\WriteBookmarks\relax
\def\floatpagepagefraction{1}
\def\textpagefraction{.001}

\shorttitle{MFE-GAN}

\shortauthors{Ju et~al.}

\title [mode = title]{MFE-GAN: Efficient GAN-based Framework for Document Image Enhancement and Binarization with Multi-scale Feature Extraction}        



\author[1]{Rui-Yang~Ju}[orcid=0000-0003-2240-1377]
\ead{jryjry1094791442@gmail.com}
\credit{Conceptualization, Methodology, Software, Validation, Data Curation, Writing – original draft, Writing - Review \& Editing}

\author[2]{~KokSheik~Wong}[orcid=0000-0002-4893-2291]
\ead{wong.koksheik@monash.edu}
\credit{Project administration, Visualization, Writing - Review \& Editing}

\author[3]{~Yanlin~Jin}[orcid=0000-0001-8466-0660]
\ead{neil.yl.jin@gmail.com}
\credit{Validation, Writing – review \& editing}

\author[4]{~Jen-Shiun~Chiang}[orcid=0000-0001-7536-8967]
\cormark[1]
\ead{jsken.chiang@gmail.com}
\credit{Resources, Funding acquisition, Supervision, Writing – review \& editing}

\affiliation[1]{
    organization={Graduate School of Informatics, Kyoto University},
    addressline={Yoshida-honmachi, Sakyo-ku}, 
    city={Kyoto},
    postcode={606-8501}, 
    country={Japan}}
\affiliation[2]{
    organization={School of Information Technology, Monash University Malaysia},
    addressline={Jalan Lagoon Selatan}, 
    city={Bandar Sunway},
    postcode={47500}, 
    country={Malaysia}}
\affiliation[3]{
    organization={Department of Electrical and Computer Engineering, Rice University},
    addressline={6100 Main St}, 
    city={Houston},
    postcode={77005}, 
    state={Texas},
    country={USA}}
\affiliation[4]{
    organization={Department of Electrical and Computer Engineering, Tamkang University},
    addressline={No.151, Yingzhuan Rd., Tamsui Dist.}, 
    city={New Taipei City},
    postcode={251301}, 
    country={Taiwan}}

\cortext[cor1]{Corresponding author: jsken.chiang@gmail.com}

\begin{abstract}
Document image enhancement and binarization are commonly performed prior to document analysis and recognition tasks for improving the efficiency and accuracy of optical character recognition (OCR) systems.
This is because directly recognizing text in degraded documents, particularly in color images, often results in unsatisfactory recognition performance.
To address these issues, existing methods train independent generative adversarial networks (GANs) for different color channels to remove shadows and noise, which, in turn, facilitates efficient text information extraction.
However, deploying multiple GANs results in long training and inference times.
To reduce both training and inference times of document image enhancement and binarization models, we propose MFE-GAN, an efficient GAN-based framework with multi-scale feature extraction (MFE), which incorporates Haar wavelet transformation (HWT) and normalization to process document images before feeding them into GANs for training.
In addition, we present novel generators, discriminators, and loss functions to improve the model's performance, and we conduct ablation studies to demonstrate their effectiveness.
Experimental results on the Benchmark, Nabuco, and CMATERdb datasets demonstrate that the proposed MFE-GAN significantly reduces the total training and inference times while maintaining comparable performance with respect to state-of-the-art (SOTA) methods.
The implementation of this work is available at \url{https://ruiyangju.github.io/MFE-GAN}.
\end{abstract}


\maketitle
\section{Introduction}
Document image enhancement and binarization are essential preprocessing steps for document analysis tasks, as they directly influence the performance of downstream tasks such as recognition and layout analysis~\cite{yang2023docdiff}.
In real-world scenarios, color-degraded documents often suffer from multiple types of degradation, including paper yellowing, text fading, and page bleeding~\cite{sun2016blind,kligler2018document,ju2025dkds}.
These degradations severely affect the image quality, thereby significantly decreasing the accuracy of optical character recognition (OCR)~\cite{mmocr2022,duan2024odm,yang2025cc} and document image understanding~\cite{kim2022ocr,wang2024docllm}.

However, for color-degraded documents, traditional image processing methods~\cite{otsu1979threshold,niblack1985introduction,sauvola2000adaptive} often fail to effectively eliminate shadows and noise, sometimes even leading to the loss of textual information.
Therefore, researchers have turned to deep learning-based methods, and many have achieved promising results.
For instance, Souibgui~\emph{et al.}~\cite{souibgui2022docentr} introduced a novel encoder-decoder architecture based on the Vision Transformer (ViT), achieving a PSNR of 19.46, an FM of 90.59, a p-FM of 93.97, and a DRD of 3.35 on the H-DIBCO~2018 dataset~\cite{pratikakis2018icdar2018}.
Yang~\emph{et al.}~\cite{yang2024gdb} proposed an end-to-end gated convolution-based network (GDB) to address the challenge of inaccurate stroke edge extraction in documents and achieved state-of-the-art (SOTA) performance on the H-DIBCO~2014 and DIBCO~2017 datasets~\cite{ntirogiannis2014icfhr2014,pratikakis2017icdar2017}.
For training and evaluation, these methods employ the ``leave-one-out'' strategy to construct the training set (viz., for the selected test set, all the remaining datasets are used to train the model).
Considering the computing resources for model training, we hypothesize that the strategy~\cite{vo2018binarization,he2019deepotsu,suh2022two,ju2023ccdwt,ju2024three} of using fixed training and test sets as the Benchmark Dataset is more efficient compared to the ``leave-one-out'' strategy.

\begin{figure*}[t]
\centering
\includegraphics[width=\textwidth]{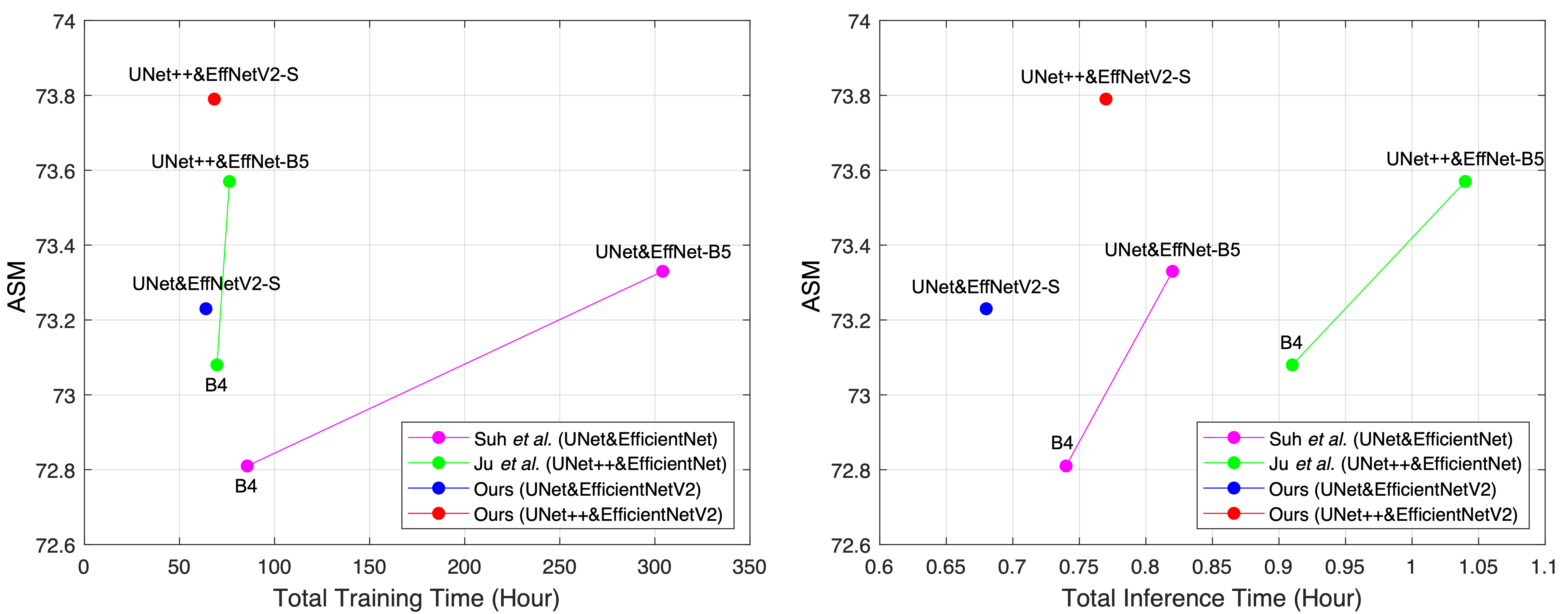}
\vskip 0.1in
\setlength{\tabcolsep}{8pt}{
\begin{tabular}{llccc}
\toprule
\textbf{Method} & \textbf{Generator} & \textbf{ASM$\uparrow$} & \textbf{Total Training Time$\downarrow$} & \textbf{Total Inference Time$\downarrow$} \\ \midrule
Suh \emph{et al.}~(PR 2022)~\cite{suh2022two} & U-Net \& EfficientNet-B5 & 73.33 & 304.12h & 0.82h \\
Ju \emph{et al.}~(KBS 2024)~\cite{ju2024three} & U-Net++ \& EfficientNet-B5 & 73.57 & 76.29h & 1.04h \\
MFE-GAN (Ours) & U-Net \& EfficientNetV2-S & 73.23 & 63.91h & 0.68h \\ 
MFE-GAN (Ours) & U-Net++ \& EfficientNetV2-S & 73.79 & 68.43h & 0.77h \\ \bottomrule
\end{tabular}}
\caption{\textbf{Graphs (top) and table (bottom)} compare the average-score metric (ASM) with respect to total training and inference times, measured on the Benchmark Dataset using an NVIDIA GeForce RTX 4090 GPU.
MFE-GAN, using U-Net \& EfficientNetV2-S as the generator, trains $16\%\sim79\%$ faster than the compared methods, while inference time is reduced by $17\%\sim35\%$.}
\label{fig:intro}
\end{figure*}

Although the existing SOTA GAN-based methods~\cite{suh2022two,ju2024three} achieve excellent performance on the Benchmark Dataset, their total training and inference times are too long due to the use of six generative adversarial networks (GANs)~\cite{goodfellow2020generative}.
As shown in Figure~\ref{fig:intro}, these computational times are prohibitively high.
To address this issue, we propose MFE-GAN, an efficient GAN-based framework that incorporates a novel multi-scale feature extraction (MFE) module, along with the generator, discriminator, and loss functions.
Furthermore, we extend our previously published conference paper~\cite{ju2025efficient} by evaluating MFE-GAN on additional datasets and providing more detailed information about our work. 
Our work makes the following contributions: 
\begin{enumerate}[(a)]
\item Incorporating both training and inference times as evaluation metrics, which were overlooked by previous methods.
\item Discovering cases where PSNR does not always accurately reflect model performance and introducing a new average score metric (ASM) for a more comprehensive evaluation.
\item Employing Haar wavelet transformation (HWT) with normalization for multi-scale feature extraction (MFE), effectively reducing training and inference time.
\item Outperforming SOTA GAN-based methods on three datasets in terms of model performance, as well as training and inference times, through the incorporation of a novel MFE module, generator, discriminator, and loss functions.
\end{enumerate}

The rest of this manuscript is organized as follows:
Section~\ref{sec:related} introduces the application of image generation networks in document image binarization and reviews SOTA methods for color document image enhancement and binarization. 
Section~\ref{sec:method} describes the proposed method, including the network architecture, multi-scale feature extraction, and loss functions.
Section~\ref{sec:experiments} analyzes the performance of the proposed method, quantitatively compares it with SOTA GAN-based methods on three datasets, and presents ablation studies to demonstrate the effectiveness of each component.
Section~\ref{sec:discussion} discusses the limitations of our method based on the analysis of visual results.
Finally, Section~\ref{sec:conclusion} concludes this work and highlights potential directions for future research.

\section{Related Work\label{sec:related}}
With the introduction of fully convolutional networks (FCNs)~\cite{long2015fully}, document image binarization has made significant progress by formulating the task as a pixel-wise prediction problem.
Tensmeyer~\emph{et al.}~\cite{tensmeyer2017document} formulated binarization as a supervised pixel classification task and demonstrated the effectiveness of FCNs for this purpose. 
Based on U-Net~\cite{ronneberger2015u}, Peng~\emph{et al.}~\cite{peng2017using} developed a convolutional encoder–decoder architecture to perform binarization. 
He~\emph{et al.}~\cite{he2019deepotsu} proposed DeepOtsu, which first employed convolutional neural networks (CNNs) for document image enhancement and then applied Otsu's method~\cite{otsu1979threshold} to produce binarized outputs. 
In addition, Zaragoza~\emph{et al.}~\cite{calvo2019selectional} employed a selective autoencoder method to parse document images, followed by global thresholding for final binarization.

The introduction of GANs~\cite{goodfellow2020generative} has further advanced image generation-based methods for document image binarization. 
For example, Zhao~\emph{et al.}~\cite{zhao2019document} formulated binarization as an image-to-image translation task, employing conditional generative adversarial networks (cGANs) to address the challenge of combining multiscale information in binarization. 
On the other hand, Souibgui~\emph{et al.}~\cite{souibgui2020gan} introduced an effective end-to-end framework based on cGANs (termed as the document enhancement generative adversarial network, DE-GAN) to restore degraded document images, achieving outstanding results on the DIBCO 2013, DIBCO 2017, and H-DIBCO 2018 datasets~\cite{pratikakis2013icdar,pratikakis2017icdar2017,pratikakis2018icdar2018}. 
Furthermore, Deng~\emph{et al.}~\cite{de2020document} proposed a method employing a dual discriminator generative adversarial network (DD-GAN) using focal loss as the generator loss function.

Recently, two SOTA document image enhancement and binarization methods have explored the use of multiple GANs for different color channels to better address color degradations.
Specifically, Suh~\emph{et al.}~\cite{suh2022two} proposed a two-stage GAN method using six improved CycleGANs~\cite{zhu2017unpaired} for color document image binarization. 
In Suh~\emph{et al.}’s model architecture, the generator adopts a U-Net~\cite{ronneberger2015u} with EfficientNet~\cite{tan2019efficientnet}, while a Pix2Pix-based conditional discriminator~\cite{isola2017image} is employed for adversarial learning.
Based on this two-stage design, Ju~\emph{et al.}~\cite{ju2024three} further introduced a three-stage GAN method, in which the same multi-GAN strategy is retained while the generator is upgraded to U-Net++~\cite{zhou2019unet++} for improved feature representation.
Although these methods consistently outperform other SOTA models on the DIBCO datasets, they suffer from unsatisfactory total training and inference times due to the use of multiple GANs. 

\begin{figure*}[t]
\centering
\includegraphics[width=\linewidth]{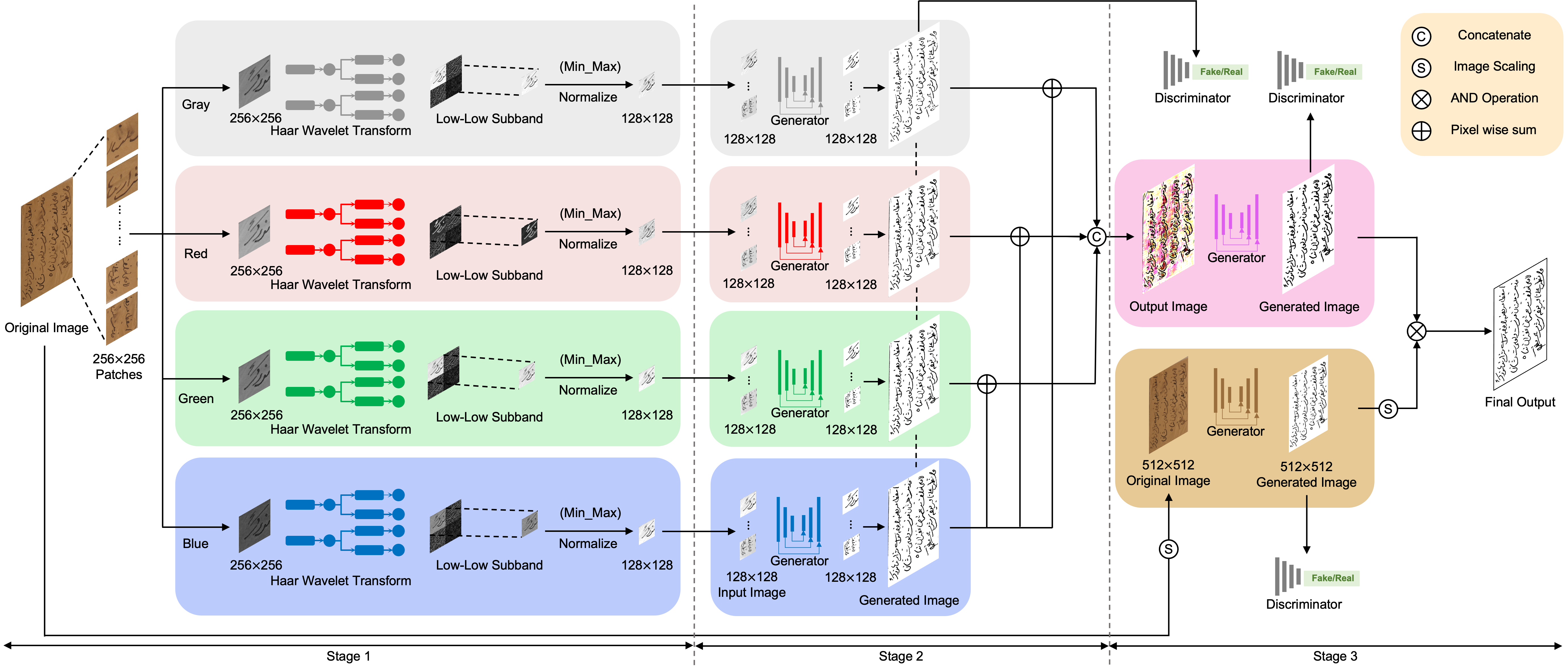}
\caption{The overall architecture of three-stage GAN-based framework, named MFE-GAN.
Stage~1: Document image processing; Stage~2: Document image enhancement; and Stage~3: Document image binarization.}
\label{fig:overall}
\end{figure*}

\section{Proposed Method\label{sec:method}}
\subsection{Network Architecture}
We propose MFE-GAN, a novel three-stage GAN-based framework to perform document image processing, enhancement, and binarization, as shown in Figure~\ref{fig:overall}. 
In Stage~1, the original color document image is divided into several 256 $\times$ 256 pixel patches. 
Each patch is then split into four single-channel images (i.e., red, green, blue, and gray), because training on separate color channels tends to yield better results.
For the MFE module, we apply HWT to each single-channel patch and extract the 128 $\times$ 128 pixel LL (low-low) sub-band. 
This sub-band is then normalized and serves as the input for Stage~2.

In Stage~2, MFE-GAN employs four independent generators using an encoder–decoder architecture based on U-Net++~\cite{zhou2019unet++} and an EfficientNetV2-S~\cite{tan2021efficientnetv2} backbone.
The effect of EfficientNet~\cite{tan2019efficientnet} and EfficientNetV2~\cite{tan2021efficientnetv2} on the MFE-GAN model is detailed in Appendix~\ref{appendix:generator}.

Then, each 128 $\times$ 128 single-channel sub-band obtained from Stage~1 is fed into its corresponding generator, which outputs a 128 $\times$ 128 enhanced sub-image.
As shown in Figure~\ref{fig:overall}, the four enhanced sub-images are first combined via pixel-wise summation and then concatenated to form the final output of Stage~2.

To ensure consistent supervision across multiple generators, a shared discriminator is used for all independent generators.
Specifically, we employ an improved PatchGAN~\cite{zhu2017unpaired} as the discriminator, applying instance normalization to all layers except the first.
This design choice avoids undesired normalization of low-level color information at the input stage, thereby preserving essential features for subsequent adversarial learning.

In Stage~3, the framework further incorporates multi-scale GANs to jointly perform local and global binarization for enhancing the distinction between text and background. 
The output of Stage~2, which maintains the same resolution as the original input image, is fed into an independent generator that produces the local binarization output ($B_{local}$).

In addition, the original input image is scaled to 512 $\times$ 512 pixels using nearest-neighbor interpolation~\cite{jahne2005digital} and fed into another independent generator to produce the global binarization output ($B_{global}$).
Each of these two branches (local and global) employs its own discriminator, forming two complete GANs.
The final output ($B_{final}$) is obtained by an AND operation of the local and global binarization results ($B_{final} = B_{local} \otimes B_{global}$).

\subsection{Multi-scale Feature Extraction}
To reduce both total training and inference times, MFE-GAN employs its multi-scale feature extraction (MFE) module on 256 $\times$ 256 pixel patches in Stage~1.
The time taken for each stage before and after using the MFE module is reported and discussed in Appendix~\ref{appendix:mfe}.

It is well known that reducing the input image size can significantly reduce the model training time, and decreasing the size of patches from $256 \times 256$ pixels to 128 $\times$ 128 pixels is consistent with this.
However, directly reducing the image size using simple interpolation would negatively impact the model's performance. 
Instead of using interpolation for image size reduction, MFE-GAN employs Haar wavelet transformation (HWT) and normalization, which effectively preserve contour information and reduce noise interference while decreasing the image size.
This approach is superior to interpolation methods that produce each output pixel based on its neighboring pixels.
We present the related experiments in Section~\ref{sec:HWT&Norm}, which demonstrate that the global binarization results of the images processed by HWT and normalization are closer to the ground-truth images than those processed by other interpolation methods.

During Stage~1 document image processing, HWT decomposes the input images into four sub-bands (LL, LH, HL, and HH). 
The low-frequency component (LL) encodes the contour information, and the high-frequency components (LH, HL, and HH) capture details and localized information. 
Therefore, we retain and normalize the low-low (LL) sub-band from HWT, effectively filtering out noise (which is often high-frequency) from color document images.

\subsection{Loss Function}
The training of generative adversarial networks (GANs) is well known to suffer from unstable loss convergence~\cite{goodfellow2020generative}.
To improve training stability in MFE-GAN, we apply the objective function of the Wasserstein Generative Adversarial Network with Gradient Penalty (WGAN-GP)~\cite{gulrajani2017improved} as the adversarial loss.

In addition to the adversarial objective, a pixel-wise supervision loss is introduced to guide the binarization process.
Since document image binarization aims to classify each pixel into one of two categories (namely, text and background), we employ binary cross-entropy (BCE) loss instead of the $L1$ loss used in the original method~\cite{isola2017image}. 
Experiments by Bartusiak~\emph{et al.}~\cite{bartusiak2019splicing} also demonstrated that BCE loss outperforms $L1$ loss in binary classification tasks.

While BCE loss focuses on the accuracy of each individual pixel, Dice loss~\cite{sudre2017generalised} emphasizes the accuracy of the entire region.
Galdran~\emph{et al.}~\mbox{\cite{galdran2022optimal}} demonstrated that combining BCE and Dice loss functions enhances segmentation performance at both the pixel-wise and region-wise levels.
Since better segmentation performance at the regional level contributes to the greater completeness of the generated text, we use an improved WGAN-GP objective loss function, which includes both BCE loss $\mathop{\mathbb{L}_{\text{BCE}}}$ and Soft Dice loss $\mathop{\mathbb{L}_{\text{Soft-DICE}}}$~\mbox{\cite{milletari2016v}}, expressed as follows:
\begin{equation}
\begin{aligned}
\mathop{\mathbb{L}_\text{G}}(x,y;\theta_G) = & -\mathop{\mathbb{E}_{x}}[D(G(x),x)] + \lambda_1 \mathop{\mathbb{L}_{\text{BCE}}}(G(x), y)\\
& + \lambda_2 \mathop{\mathbb{L}_{\text{Soft-DICE}}}(G(x), y),
\label{eq:theta_G}
\end{aligned}
\end{equation}
\begin{equation}
\begin{aligned}
\mathbb{L}_\text{D} = &-\mathbb{E}_{x,y}[D(y,x)] + \mathbb{E}_{x}[D(G(x), x)] \\
&+ \alpha \mathbb{E}_{x, \hat{y}\sim P_{\hat{y}}}[(\lVert \nabla_{\hat{y}} D(\hat{y}, x) \lVert_2 - 1 )^2],
\label{eq:theta_D}
\end{aligned}
\end{equation}
where $x$ is the input image, $G(x)$ is the generated image, and $y$ is the ground-truth image. 
Here, $\lambda_1$ and $\lambda_2$ control the relative importance of different loss terms, while $\alpha$ denotes the gradient penalty coefficient. 
The discriminator $D$ is trained for minimizing $\mathbb{L}_\text{D}$ to distinguish between ground-truth and generated images, while the generator $G$ aims to minimize $\mathbb{L}_\text{G}$.
The equations for BCE loss $\mathop{\mathbb{L}_{\text{BCE}}}$ and Soft Dice loss $\mathop{\mathbb{L}_{\text{Soft-DICE}}}$ are shown as follows:
\begin{equation}
\begin{aligned}
\mathop{\mathbb{L}_{\text{BCE}}}(\mathbf{\hat{y}},\mathbf{y})= \mathop{\mathbb{E}_{\mathbf{\hat{y}},\mathbf{y}}} [\mathbf{y}\log \mathbf{\hat{y}} + (1-\mathbf{y})\log (1-\mathbf{\hat{y}})],
\end{aligned}
\label{eq:bce_loss}
\end{equation}
\begin{equation}
\begin{aligned}
\mathop{\mathbb{L}_{\text{Soft-DICE}}}(\mathbf{\hat{y}},\mathbf{y})= 1 - \frac{2|\mathbf{\hat{y}} \cap \mathbf{y}|}{|\mathbf{\hat{y}}| + |\mathbf{y}|} = 1-\frac{2\langle \mathbf{\hat{y}}, \mathbf{y}\rangle}{\langle \mathbf{\hat{y}}, \mathbf{\hat{y}}\rangle + \langle \mathbf{y}, \mathbf{y}\rangle},
\end{aligned}
\label{eq:dice_loss}
\end{equation}
where $\mathbf{y}$ is the ground-truth, and $\mathbf{\hat{y}}$ is the predicted image.
The performance of models trained with different loss function configurations is reported in Appendix~\ref{appendix:loss}.

\begin{table*}[t]
\centering
\caption{Detailed utilization statistics for the three datasets considered in this work.} 
\setlength{\tabcolsep}{8pt}{
\begin{tabular}{ccccccc}
\toprule
\multirow{2}{*}{\textbf{Dataset}} & \multirow{2}{*}{\textbf{Strategy}} & \multicolumn{3}{c}{\textbf{Training Set (Pages)}} & \multicolumn{2}{c}{\textbf{Test Set (Pages)}} \\ \cmidrule(r){3-5} \cmidrule(r){6-7}
 &  & Original & Processed ($256^2$) & Processed ($512^2$) & Original & Processed ($512^2$) \\ \midrule
Benchmark$^1$ & Following \cite{suh2022two,ju2024three} & 143 & 120,174 & 804 & 102 & 582 \\ \midrule
\multirow{2}{*}{Nabuco} & \multirow{2}{*}{\begin{tabular}[c]{@{}c@{}}Two-Fold\\Cross-Validation\end{tabular}} & 15 & 32,038 & 90 & 20 & 120 \\
 &  & 20 & 48,400 & 120 & 15 & 90 \\ \midrule
\multirow{5}{*}{CMATERdb} & \multirow{5}{*}{\begin{tabular}[c]{@{}c@{}}Five-Fold\\Cross-Validation\end{tabular}} & 4 & 5,308 & 24 & 1 & 6 \\ 
 &  & 4 & 5,242 & 24 & 1 & 6 \\ 
 &  & 4 & 4,942 & 24 & 1 & 6 \\ 
 &  & 4 & 4,592 & 24 & 1 & 6 \\ 
 &  & 4 & 2,140 & 24 & 1 & 6 \\ \bottomrule
\end{tabular}}
\begin{minipage}{0.95\textwidth}
\footnotesize
$^1$ ``Benchmark'' refers to training sets containing DIBCO 2009, H-DIBCO 2010, H-DIBCO 2012, BD, PHIBD, and SMADI; 
and test sets containing DIBCO 2011, DIBCO 2013, H-DIBCO 2014, H-DIBCO 2016, DIBCO 2017, H-DIBCO 2018, and DIBCO 2019.
\end{minipage}
\label{tab:dataset}
\end{table*}

\begin{figure*}[t]
\centering
\includegraphics[width=\linewidth]{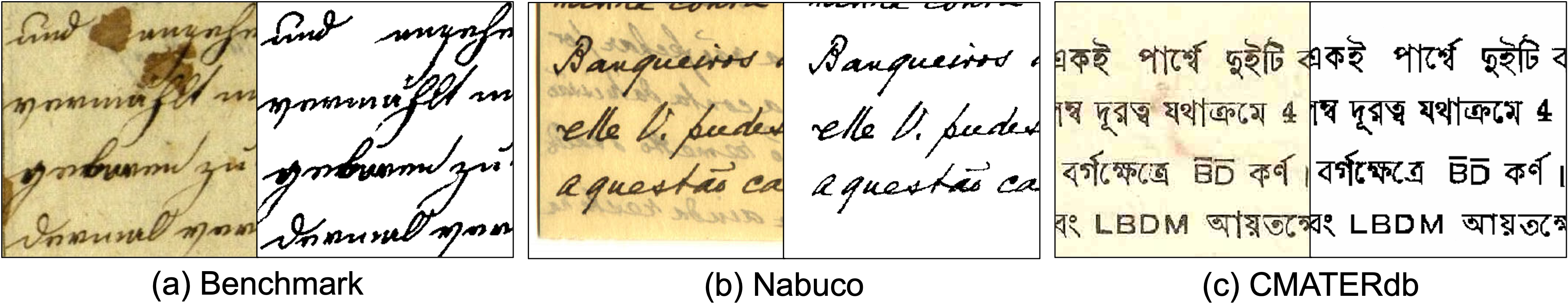}
\caption{Representative examples from the three datasets used in this work: (a) Benchmark, (b) Nabuco, and (c) CMATERdb. 
Original images are shown on the left, and their corresponding binarized ground-truth images are shown on the right.}
\label{fig:dataset}
\end{figure*}

\begin{table}[t]
\centering
\caption{Comparison of baseline and proposed model configurations.}
\setlength{\tabcolsep}{8pt}{
\begin{tabular}{|c|c|}
\hline 
\multicolumn{2}{|c|}{\textbf{Baseline~\cite{suh2022two}}} \\ \hline 
\textbf{Generator} & U-Net \& EfficientNet-B5 \\ \hline 
\textbf{Generator Loss} & $- D(G(z)) + \lambda \mathbb{L}_{\text{BCE}}$ \\ \hline
\textbf{Discriminator} & Similar Pix2pixGAN \\ \hline
\textbf{Discriminator Loss} & $D(x) - D(G(z)) + 10 \cdot \text{GP}$ \\ \hline 
\textbf{MFE Module} & \ding{55} \\ \hline 
\noalign{\smallskip} \noalign{\smallskip} \noalign{\smallskip} \hline
\multicolumn{2}{|c|}{\textbf{MFE-GAN}} \\ \hline
\textbf{Generator} & U-Net++ \& EfficientNetV2-S \\ \hline 
\textbf{Generator Loss} & $- D(G(z)) + \lambda_1 \mathbb{L}_{\text{BCE}} + \lambda_2 \mathbb{L}_{\text{Soft-DICE}}$ \\ \hline
\textbf{Discriminator} & Improved PatchGAN \\ \hline
\textbf{Discriminator Loss} & $D(x) - D(G(z)) + 10 \cdot \text{GP}$ \\ \hline
\textbf{MFE Module} & \ding{51} \\ \hline \noalign{\smallskip}
\end{tabular}}
\begin{minipage}{0.95\linewidth}
\footnotesize
MFE: Multi-scale Feature Extraction; GP: Gradient Penalty; $\lambda = 50$, and $\lambda_1 = \lambda_2 = 25$.
\end{minipage}
\label{tab:baseline}
\end{table}

\begin{table*}[t]
\centering
\caption{PSNR (dB) of images resized using different methods: Interpolation/HWT/HWT\&Normalization (Ours), for various datasets.}
\setlength{\tabcolsep}{7.5pt}{
\begin{tabular}{lccccccc}
\toprule
\textbf{Method} & \textbf{DIBCO 2009} & \textbf{H-DIBCO 2010} & \textbf{H-DIBCO 2012} & \textbf{Bickley Diary} & \textbf{PHIBD} & \textbf{SMADI} & \textbf{Mean Values} \\ \cmidrule(r){1-1} \cmidrule(r){2-7} \cmidrule(r){8-8}
Bicubic & {\color{blue}71.45dB} & {\color{blue}72.22dB} & 71.67dB & 64.29dB & 69.58dB & {\color{blue}69.88dB} & {\color{blue}69.85dB} \\
Bilinear & 70.94dB & 72.16dB & 71.46dB & 64.07dB & {\color{blue}69.71dB} & 69.86dB & 69.70dB \\
Area & 70.94dB & 72.16dB & 71.46dB & 64.07dB & {\color{blue}69.71dB} & 69.86dB & 69.70dB \\
Nearest & 70.95dB & 72.04dB & 71.59dB & 64.20dB & 69.69dB & 69.83dB & 69.72dB \\
Lanczos & 71.42dB & {\color{blue}72.22dB} & {\color{blue}71.69dB} & {\color{blue}64.30dB} & 69.58dB & {\color{red}69.89dB} & {\color{blue}69.85dB} \\
HWT & 62.65dB & 67.11dB & 59.67dB & 53.76dB & 58.00dB & 59.48dB & 60.11dB \\
\textbf{Ours} & {\color{red}71.77dB} & {\color{red}72.74dB} & {\color{red}72.85dB} & {\color{red}64.44dB} & {\color{red}70.76dB} & 69.44dB & {\color{red}70.34dB} \\ \bottomrule
\end{tabular}}
\begin{minipage}{0.95\textwidth}
\footnotesize
Our method uses HWT\&Normalization.
The highest and second-highest PSNR values are highlighted in {\color{red}red} and {\color{blue}blue}, respectively.
\end{minipage}
\label{tab:psnr}
\end{table*}

\section{Experiments\label{sec:experiments}}
\subsection{Datasets\label{sec:datasets}}
\subsubsection{Benchmark Dataset}
\textbf{DIBCO} (Document Image Binarisation Contest) provides ten competition datasets, including DIBCO 2009 \cite{gatos2009icdar}, H-DIBCO 2010 \cite{pratikakis2010h}, DIBCO 2011 \cite{pratikakis2011icdar2011}, H-DIBCO 2012 \cite{pratikakis2012icfhr}, DIBCO 2013 \cite{pratikakis2013icdar}, H-DIBCO 2014 \cite{ntirogiannis2014icfhr2014}, H-DIBCO 2016 \cite{pratikakis2016icfhr2016}, DIBCO 2017 \cite{pratikakis2017icdar2017}, H-DIBCO 2018 \cite{pratikakis2018icdar2018}, and DIBCO 2019 \cite{pratikakis2019icdar2019}. 
These datasets include both machine-printed and handwritten images in grayscale and color, with a total of 136 images.

\textbf{BD} (Bickley Diary)~\cite{deng2010binarizationshop} was generously donated to the Singapore Methodist Archives by Mr. Erin Bickley. 
This dataset contains seven diary images, where factors such as lighting variations and fold damage make text recognition particularly challenging.

\textbf{PHIBD} (Persian Heritage Image Binarization Dataset) \cite{nafchi2013efficient} comprises 15 historical manuscript images sourced from Mr. Mirza Mohammad Kazemaini's old manuscript collection in Yazd, Iran. 
The manuscripts within the images are affected by various degrees of degradation, including bleed-through, fading, and blurring.

\textbf{SMADI} (Synchromedia Multispectral Ancient Document Images)~\cite{hedjam2013historical} was captured using a CROMA CX MSI camera, producing eight images for each document, resulting in a total of 240 images of authentic documents. 
The ancient documents in these images were written in iron-bile ink and date from the 17th to 20th centuries. 

To ensure a fair comparison between the proposed MFE-GAN and the SOTA GAN-based methods~\cite{suh2022two,ju2024three}, we adopt the same strategy as in~\cite{suh2022two,ju2024three} to construct the training set, as detailed in Table~\ref{tab:dataset}.
The training set comprises images from DIBCO 2009 (10 images); H-DIBCO 2010 (10 images); H-DIBCO 2012 (14 images); Bickley Diary (7 images); PHIBD (15 images); and SMADI (87 images).
The testing set consists of images from DIBCO 2011 (16 images); DIBCO 2013 (16 images); H-DIBCO 2014 (10 images); H-DIBCO 2016 (10 images); DIBCO 2017 (20 images); H-DIBCO 2018 (10 images); and DIBCO 2019 (20 images).
Representative examples from the Benchmark Dataset are shown in Figure~\ref{fig:dataset}.

\subsubsection{Nabuco Dataset}
\textbf{Nabuco}~\cite{lins2011nabuco} images were digitally compiled by Rafael Dueire Lins and historians from the Joaquim Nabuco Foundation between 1992 and 1994, using a true-color table scanner with a resolution of 200 dpi. 
The Nabuco bequest, consisting of 6,500 letters and postcards, both handwritten and typed, comprises approximately 30,000 pages.
This bequest holds significant value for research on the history of the Americas, as Joaquim Nabuco was one of the key figures in the abolition of slavery and the first Brazilian ambassador to the United States.

Due to the lack of corresponding ground-truth for most Nabuco images, we use only 35 images with available ground-truth, provided by the DIB team at CIn-UFPE, Brazil.
Representative examples from these images are presented in Figure~\ref{fig:dataset}.
This team offers two datasets, containing 15 and 20 color Nabuco images with their respective ground-truth. 
For evaluation, we perform a two-fold cross-validation procedure on the Nabuco dataset.
The specific implementation details are shown in Table~\ref{tab:dataset}.

\subsubsection{CMATERdb Dataset}
\textbf{CMATERdb}~\cite{mollah2012computationally} is a dataset of Bengali and English manuscripts created by the Center for Microprocessor Applications for Training Education and Research (CMATER) at Jadavpur University, India. 
It comprises 5 images of color documents, including both camera-captured and scanned materials. 
These 5 images include a diverse range of document types, such as historical manuscripts and contemporary records, as well as degraded and well-preserved documents. 
Examples from this dataset are shown in Figure~\ref{fig:dataset}. 

Since this dataset consists of only 5 images, each representing different conditions (i.e., blurred vs. clear, degraded vs. well-preserved), we perform a five-fold cross-validation procedure on the CMATERdb dataset.
Specifically, we select four images for training and one for testing. 
The detailed information is provided in Table~\ref{tab:dataset}.

\subsection{Evaluation Metrics}
For quantitative comparison, four classical evaluation metrics are employed, namely: the f-measure (FM), pseudo-f-measure (p-FM), peak signal-to-noise ratio (PSNR), and distance reciprocal distortion (DRD). 
Although these metrics jointly characterize binarization quality from different perspectives, inconsistencies may arise when comparing different methods.
For instance, the proposed MFE-GAN achieves SOTA-level FM and p-FM values, but its PSNR is lower than that of other methods, making it difficult to assess overall performance based on any single metric alone.

To address this issue, and inspired by Jemni~\mbox{\emph{et al.}~\cite{jemni2022enhance}}, we introduce the average-score metric (ASM) to evaluate the overall performance of each method more comprehensively:
\begin{equation}
ASM = \frac{FM + p\text{-}FM + PSNR + (100 - DRD)}{4}.
\end{equation}

Note that in ASM, segmentation-quality metrics (FM, p-FM) are balanced against pixel-wise metrics (PSNR, DRD). 
We consider this reasonable, as it prevents a single metric, such as a low PSNR, from disproportionately penalizing an otherwise effective model.
This is because, for methods utilizing GANs to generate binarized images, the focus should be on the overall quality of the generated binarized image rather than on individual pixels.
Furthermore, as we illustrate in Section~\ref{sec:visual}, our proposed MFE-GAN can generate more complete images, although its PSNR is lower than that of the compared methods.

In addition, to demonstrate the efficiency of our proposed MFE-GAN relative to others when using the same computational resources, we calculate the total training and inference times for all models in hours (h). 
The total training time includes: the training time of models in Stage~2, the time required for Stage~2 models to generate all output images (Stage~2 Predict), the training time for the model using Stage~2's output images in Stage~3 (Stage~3 Top), and the training time for the model using resized original images ($512 \times 512$) in Stage~3 (Stage~3 Bottom). 
The total inference time is defined as the time required to generate all images of the test set.

\subsection{Baseline}
We select the method~\cite{suh2022two}, which has a similar network architecture, as the baseline.
As shown in Table~\ref{tab:baseline}, the baseline method differs from MFE-GAN in terms of the generator and discriminator, the loss functions, and the multi-scale feature extraction (MFE) module.

\subsection{Implementation Details}
\subsubsection{Data Preparation\label{subsec:data}}
To ensure a fair comparison, we employ the same dataset and data augmentation strategies for both MFE-GAN and the SOTA GAN-based methods~\cite{suh2022two,ju2024three}.
In Stage~1, the original input images are split into 256 $\times$ 256 pixel patches to match the input size of the ImageNet~\cite{deng2009imagenet} dataset, as we utilize a pre-trained model based on this dataset.
Data augmentation is applied to expand the training samples, with scaling factors of 0.75, 1, 1.25, and 1.5, as well as a rotation of 270$\degree$.
For the Benchmark Dataset, this results in a total of 120,174 training image patches.

For global binarization in Stage~3, the input images are resized to 512 $\times$ 512 pixels.
This set is further augmented through horizontal and vertical flipping, as well as rotations of 90$\degree$, 180$\degree$, and 270$\degree$, resulting in 804 training images of size 512 $\times$ 512 pixels for the Benchmark Dataset.

The Nabuco and CMATERdb datasets employ the same data augmentation strategies for their respective stages. 
The final number of processed training image patches and resized 512 $\times$ 512 pixel training images for these datasets is summarized in Table~\ref{tab:dataset}.

\subsubsection{Pre-training and Training}
All methods employ pre-trained weights from the ImageNet~\cite{deng2009imagenet} dataset to improve training efficiency, due to constraints on data availability. 
Specifically, Suh~\emph{et al.}~\cite{suh2022two} and Ju~\emph{et al.}~\cite{ju2024three} use EfficientNet~\cite{tan2019efficientnet} as the encoder of their GANs, while MFE-GAN adopts EfficientNetV2~\cite{tan2021efficientnetv2}.

\subsubsection{Training}
To ensure a fair comparison of training and inference times, all models are trained on NVIDIA RTX 4090 GPUs using PyTorch as the implementation framework. 
The training parameters for Stage~2 and Stage~3 are largely similar, with the main exception being the number of epochs: 10 for Stage~2 and 150 for Stage~3. 
We choose the Adam optimizer to train the models and set the initial learning rate to 2$\times10^{-4}$. 
In addition, the Adam optimizer parameters are set to $\beta_1 = 0.5$ and $\beta_2 = 0.999$ for training both the generators and the discriminators.
Additional implementation details and training scripts are available in our GitHub repository for reproducibility.

\begin{table*}[t]
\centering
\caption{Training and inference time (hours) of the proposed and SOTA GAN-based methods on the \textbf{Benchmark Dataset}.}
\setlength{\tabcolsep}{13pt}{
\begin{tabular}{lcccccc}
\toprule
\textbf{Method} & \textbf{\begin{tabular}[c]{@{}c@{}}Stage2 \\ Train\end{tabular}} & \textbf{\begin{tabular}[c]{@{}c@{}}Stage2 \\ Predict\end{tabular}} & \textbf{\begin{tabular}[c]{@{}c@{}}Stage3 \\ Top\end{tabular}} & \textbf{\begin{tabular}[c]{@{}c@{}}Stage3 \\ Bottom\end{tabular}} & \textbf{\begin{tabular}[c]{@{}c@{}}Total\\ Training\end{tabular}} & \textbf{\begin{tabular}[c]{@{}c@{}}Total\\ Inference\end{tabular}} \\ \cmidrule(r){1-1} \cmidrule(r){2-5} \cmidrule(r){6-7}
U-Net \& EfficientNet-B4~\cite{suh2022two} & 14.73h & 3.75h & 65.96h & {\color{red}1.17h} & 85.61h & {\color{blue}0.74h} \\ 
U-Net \& EfficientNet-B5~\cite{suh2022two} & 16.30h & 3.77h & 282.80h & {\color{blue}1.26h} & 304.12h & 0.82h \\
U-Net++ \& EfficientNet-B4~\cite{ju2024three} & 18.81h & 3.95h & {\color{red}45.63h} & 1.29h & 69.68h & 0.91h \\ 
U-Net++ \& EfficientNet-B5~\cite{ju2024three} & 21.23h & 4.37h & 49.23h & 1.46h & 76.29h & 1.04h \\
\textbf{U-Net \& EfficientNetV2-S} & {\color{red}11.60h} & {\color{red}3.45h} & {\color{blue}47.47h} & 1.39h & {\color{red}63.91h} & {\color{red}0.68h} \\
\textbf{U-Net++ \& EfficientNetV2-S} & {\color{blue}14.12h} & {\color{blue}3.63h} & 49.29h & 1.39h & {\color{blue}68.43h} & 0.77h \\ \bottomrule
\end{tabular}}
\begin{minipage}{0.95\textwidth}
\footnotesize
The proposed MFE-GAN is highlighted in \textbf{bold}. 
The best and second-best performances are colored in {\color{red}red} and {\color{blue}blue}, respectively.
\end{minipage}
\label{tab:comparison_time}
\end{table*}

\begin{table*}[t]
\centering
\caption{Quantitative comparison (ASM: FM/p-FM/PSNR/DRD, Total Training Time, Total Inference Time) of the proposed MFE-GAN and the considered methods for document image enhancement and binarization on the \textbf{Benchmark Dataset}.}
\setlength{\tabcolsep}{10pt}{
\begin{tabular}{lccccccc}
\toprule
\textbf{Method} & \textbf{FM$\uparrow$} & \textbf{p-FM$\uparrow$} & \textbf{PSNR$\uparrow$} & \textbf{DRD$\downarrow$} & \textbf{ASM$\uparrow$} & \textbf{Training$\downarrow$} & \textbf{Inference$\downarrow$} \\ \cmidrule(r){1-1} \cmidrule(r){2-5} \cmidrule(r){6-8}
Otsu~\cite{otsu1979threshold} & 73.91 & 75.93 & 14.50dB & 30.32 & 58.51 & -- & -- \\
Sauvola~\cite{sauvola2000adaptive} & 75.83 & 80.72 & 15.62dB & 9.65 & 65.63 & -- & -- \\
U-Net \& EfficientNet-B4~\cite{suh2022two} & 87.95 & 89.01 & 19.10dB & 4.83 & 72.81 & 85.61h & {\color{blue}0.74h} \\
U-Net \& EfficientNet-B5~\cite{suh2022two} & 88.56 & 89.90 & {\color{red}19.31dB} & {\color{red}4.46} & 73.33 & 304.12h & 0.82h \\
U-Net++ \& EfficientNet-B4~\cite{ju2024three} & 88.14 & 89.71 & 19.09dB & 4.64 & 73.08 & 69.68h & 0.91h \\
U-Net++ \& EfficientNet-B5~\cite{ju2024three} & {\color{blue}89.13} & {\color{blue}90.35} & {\color{blue}19.30dB} & 4.49 & {\color{blue}73.57} & 76.29h & 1.04h \\ 
\textbf{U-Net \& EfficientNetV2-S} & 88.83 & 89.87 & 19.07dB & 4.86 & 73.23 & {\color{red}63.91h} & {\color{red}0.68h} \\
\textbf{U-Net++ \& EfficientNetV2-S} & {\color{red}89.69} & {\color{red}90.78} & 19.15dB & {\color{blue}4.45} & {\color{red}73.79} & {\color{blue}68.43h} & 0.77h \\ \bottomrule
\end{tabular}}
\begin{minipage}{0.95\textwidth}
\footnotesize
The proposed MFE-GAN is highlighted in \textbf{bold}. 
The best and second-best performances are colored in {\color{red}red} and {\color{blue}blue}, respectively.
\end{minipage}
\label{tab:comparison_performance}
\end{table*}

\subsection{Quantitative Comparisons}
\subsubsection{Multi-scale Feature Extraction\label{sec:HWT&Norm}}
We explore other image-resizing techniques for multi-scale feature extraction, including interpolation-based algorithms such as bicubic, bilinear, area, nearest-neighbor, and Lanczos. 
We implement these techniques using the open-source computer vision library (OpenCV) to downscale all input images and the corresponding ground-truth images from 256 $\times$ 256 to 128 $\times$ 128.
In addition, we employ the ``HWT'' and ``HWT and normalization'', i.e., our proposed MFE module. 
Note that the resized images obtained from all these methods are non-binary.

To conduct a meaningful, apples-to-apples comparison against the binary ground-truth images, we must first binarize these intermediate outputs.
Therefore, we apply a standard global thresholding algorithm (Otsu's method~\cite{otsu1979threshold}) to these resized images, and then compute the PSNR values. 
We evaluate the impact of different image resizing techniques on six training sets by calculating the PSNR values (against the corresponding ground-truth images) and computing the mean PSNR value for all images. 
The results are recorded in Table~\ref{tab:psnr}. 

We observe that the mean PSNR value achieved by the ``HWT'' method is 60.11dB, indicating that images reduced directly using HWT have low similarity with the corresponding ground-truth images. 
In addition, the mean PSNR values for resized images produced by different interpolation methods are all below 70dB. 
However, the mean PSNR value for images processed by ``HWT and normalization'' reaches 70.34dB, which confirms that the images obtained by this method are closer to the corresponding ground-truth images at the pixel level.
In conclusion, the results demonstrate that our ``HWT and normalization'' method is more effective than other interpolation-based image-resizing techniques for document image enhancement and binarization.

\subsubsection{Benchmark Dataset}
We compare MFE-GAN with the SOTA GAN-based methods~\cite{suh2022two,ju2024three} for document image enhancement and binarization.
As shown in Table~\ref{tab:comparison_time}, the compared methods using U-Net \& EfficientNet-B5 or U-Net++ \& EfficientNet-B5~\cite{tan2019efficientnet} are already slower than our proposed MFE-GAN.
Therefore, we do not further compare against methods using EfficientNet-B6, as this would further contradict our goal of reducing training and inference times.
For completion of discussion, detailed performance comparisons of different models on each test set of the Benchmark Dataset are provided in Appendix~\ref{appendix:dibco}.

Table~\ref{tab:comparison_performance} shows that MFE-GAN using U-Net++~\cite{zhou2019unet++} with EfficientNetV2-S~\cite{tan2021efficientnetv2} achieves the highest ASM of 73.79. 
It requires a total training time of 68.43h, which is also the second-shortest time.
This is faster than the U-Net++ \& EfficientNet-B5 method (76.29h), which yielded the second-highest ASM.
Furthermore, the total inference time of MFE-GAN is 0.77h, notably lower than the 1.04h required by Ju~\emph{et al.}'s method~\cite{ju2024three} (using U-Net++ \& EfficientNet-B5), representing a reduction of approximately 26\%.

In addition, MFE-GAN using U-Net \& EfficientNetV2-S obtains the shortest total training and inference times (63.91h and 0.68h, respectively).
Although the ASM value achieved by MFE-GAN of 73.23 is not the highest, when compared to Suh~\emph{et al.}'s method~\cite{suh2022two} (using U-Net with EfficientNet-B5) that yields 73.33 ASM, the training time is reduced from 304.12h to 63.91h, which is a remarkable decrease of approximately 78\%.
Overall, the experimental results demonstrate the efficiency and competitive performance of our proposed MFE-GAN.

Next, we compare the results achieved by all benchmark methods for each evaluation metric. 
MFE-GAN achieves the highest FM and p-FM values of 89.69 and 90.78, respectively, while maintaining lower total training and inference times than the method with the second highest FM and p-FM values. 
For the DRD metric, MFE-GAN achieves the second-highest value, but with a significantly reduced total training time of 68.43h compared to the 304.12h taken by the method with the highest DRD value.
Although MFE-GAN does not achieve the highest PSNR, this metric does not directly reflect model performance in document image enhancement and binarization, as we discuss in detail in Section~\ref{sec:visual}.

\begin{table*}[t]
\centering
\caption{Quantitative comparison (ASM: FM/p-FM/PSNR/DRD, Total Training Time, Total Inference Time) of the proposed MFE-GAN and SOTA GAN-based methods for document image enhancement and binarization on the \textbf{Nabuco dataset}.}
\setlength{\tabcolsep}{3pt}{
\begin{tabular}{lcccccc}
\toprule
\textbf{Method} & Suh \emph{et al.}~\cite{suh2022two} & Suh \emph{et al.}~\cite{suh2022two} & Ju \emph{et al.}~\cite{ju2024three} & Ju \emph{et al.}~\cite{ju2024three} & \textbf{MFE-GAN} & \textbf{MFE-GAN} \\
\textbf{Generator} & U-Net \& B4 & U-Net \& B5 & U-Net++ \& B4 & U-Net++ \& B5 & \textbf{U-Net \& V2-S} & \textbf{U-Net++ \& V2-S} \\ \midrule
\textbf{FM$\uparrow$} & 85.93 & 87.45 & 85.95 & {\color{blue}87.63} & 87.56 & {\color{red}88.04} \\
\textbf{p-FM$\uparrow$} & 86.57 & 88.16 & 86.37 & {\color{blue}88.27} & 88.22 & {\color{red}88.72} \\
\textbf{PSNR$\uparrow$} & 18.17dB & {\color{blue}18.61dB} & 18.17dB & {\color{red}18.65dB} & 18.51dB & 18.60dB \\
\textbf{DRD$\downarrow$} & 6.33 & 5.40 & 5.69 & 5.17 & {\color{blue}5.10} & {\color{red}5.06} \\ \midrule
\textbf{Stage2 Train Time$\downarrow$} & 4.56h & 5.71h & 5.69h & 6.62h & {\color{red}2.94h} & {\color{blue}3.02h} \\
\textbf{Stage2 Predict Time$\downarrow$} & {\color{red}2.04h} & {\color{blue}2.10h} & 2.22h & 4.77h & 2.19h & 2.22h \\
\textbf{Stage3 Top Time$\downarrow$} & 20.36h & 91.71h & 22.20h & 25.58h & {\color{red}18.30h} & {\color{blue}20.14h} \\
\textbf{Stage3 Bottom Time$\downarrow$} & {\color{red}0.51h} & {\color{red}0.51h} & {\color{blue}0.52h} & 0.54h & 0.55h & 0.56h \\ \midrule
\textbf{ASM$\uparrow$} & 71.08 & 72.21 & 71.20 & {\color{blue}72.34} & 72.30 & {\color{red}72.58} \\ 
\textbf{Total Train Time$\downarrow$} & 27.47h & 100.03h & 30.63h & 37.51h & {\color{red}23.97h} & {\color{blue}25.93h} \\
\textbf{Total Inference Time$\downarrow$} & {\color{blue}0.19h} & 0.22h & 0.23h & 0.26h & {\color{red}0.18h} & 0.21h \\ \bottomrule
\end{tabular}}
\begin{minipage}{0.95\textwidth}
\footnotesize
The best and second-best performances are highlighted in {\color{red}red} and {\color{blue}blue}, respectively.
\end{minipage}
\label{tab:nabuco}
\end{table*}

\begin{table*}[t]
\centering
\caption{Quantitative comparison (ASM: FM/p-FM/PSNR/DRD, Total Training Time, Total Inference Time) of the proposed MFE-GAN and the considered methods for document image enhancement and binarization on the \textbf{CMATERdb Dataset}.}
\setlength{\tabcolsep}{3pt}{
\begin{tabular}{lcccccc}
\toprule
\textbf{Method} & Suh \emph{et al.}~\cite{suh2022two} & Suh \emph{et al.}~\cite{suh2022two} & Ju \emph{et al.}~\cite{ju2024three} & Ju \emph{et al.}~\cite{ju2024three} & \textbf{MFE-GAN} & \textbf{MFE-GAN} \\
\textbf{Generator} & U-Net \& B4 & U-Net \& B5 & U-Net++ \& B4 & U-Net++ \& B5 & \textbf{U-Net \& V2-S} & \textbf{U-Net++ \& V2-S} \\ \midrule
\textbf{FM$\uparrow$} & 82.19 & 83.10 & 87.06 & {\color{blue}87.24} & 87.22 & {\color{red}87.36} \\
\textbf{p-FM$\uparrow$} & 88.17 & 89.44 & 91.49 & {\color{blue}92.31} & 91.66 & {\color{red}92.46} \\
\textbf{PSNR$\uparrow$} & 16.37dB & 16.85dB & 17.76dB & {\color{blue}17.83dB} & 17.80dB & {\color{red}17.85dB} \\
\textbf{DRD$\downarrow$} & 6.36 & 5.59 & 4.34 & {\color{blue}4.24} & 4.29 & {\color{red}4.19} \\ \midrule
\textbf{Stage2 Train Time$\downarrow$} & 1543.30s & 1878.08s & 2023.23s & 2276.39s & {\color{red}540.16s} & {\color{blue}609.36s} \\
\textbf{Stage2 Predict Time$\downarrow$} & {\color{red}455.98s} & 559.37s & {\color{blue}541.80s} & 553.17s & 554.10s & 556.45s \\
\textbf{Stage3 Top Time$\downarrow$} & 7974.51s & 42333.73s & {\color{red}6810.23s} & 7392.94s & {\color{blue}7213.80s} & 7533.96s \\
\textbf{Stage3 Bottom Time$\downarrow$} & 593.68s & {\color{red}554.37s} & {\color{blue}573.99s} & 579.52s & 585.98s & 599.41s \\ \midrule
\textbf{ASM$\uparrow$} & 70.09 & 70.95 & 72.99 & {\color{blue}73.28} & 73.10 & {\color{red}73.37} \\
\textbf{Total Train Time$\downarrow$} & 10567.47s & 45325.55s & 9949.25s & 10802.02s & {\color{red}8894.04s} & {\color{blue}9299.18s} \\
\textbf{Total Inference Time$\downarrow$} & 3.49s & 3.87s & 4.07s & 4.65s & {\color{red}3.06s} & {\color{blue}3.42s} \\ \bottomrule 
\end{tabular}}
\begin{minipage}{0.95\textwidth}
\footnotesize
The best and second-best performances are highlighted in {\color{red}red} and {\color{blue}blue}, respectively.
\end{minipage}
\label{tab:cmaterdb}
\end{table*}

\begin{figure*}[t]
\centering
\includegraphics[width=\linewidth]{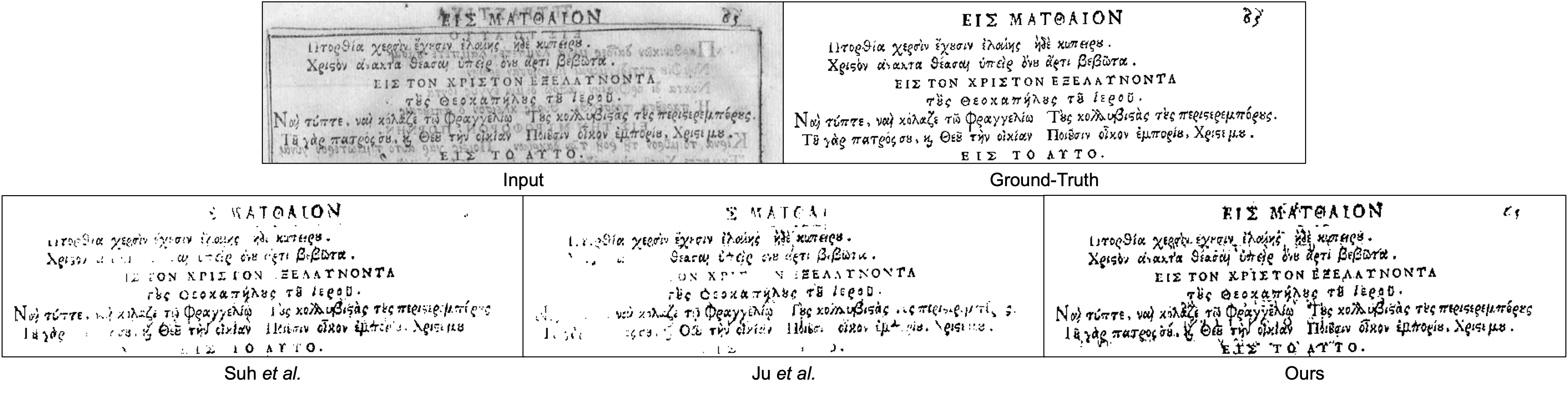}
\vskip 0.01in
\setlength{\tabcolsep}{18pt}{
\begin{tabular}{llcccc}
\toprule
\textbf{Method} & \textbf{Generator} & \textbf{FM$\uparrow$} & \textbf{p-FM$\uparrow$} & \textbf{PSNR$\uparrow$} & \textbf{DRD$\downarrow$} \\ \midrule
Blank Image & -- & -- & -- & 11.09dB & 5.11 \\
Suh \emph{et al.}~\cite{suh2022two} & U-Net \& EfficientNet-B4 & 76.71 & 76.87 & 14.83dB & 2.44 \\
Ju \emph{et al.}~\cite{ju2024three} & U-Net \& EfficientNet-B4 & 68.15 & 69.12 & 13.72dB & 3.22 \\
MFE-GAN (Ours) & U-Net \& EfficientNetV2-S & 82.42 & 82.27 & 14.96dB & 3.45 \\ \bottomrule
\end{tabular}}
\caption{Representative visualized results from the test set (case 1): the first row from left to right shows input image and ground-truth image; the second row from left to right shows Suh~\emph{et al.}~\cite{suh2022two}, Ju~\emph{et al.}~\cite{ju2024three}, and MFE-GAN (Ours).} 
\label{fig:result1}
\end{figure*}

\begin{figure*}[t]
\centering
\includegraphics[width=\linewidth]{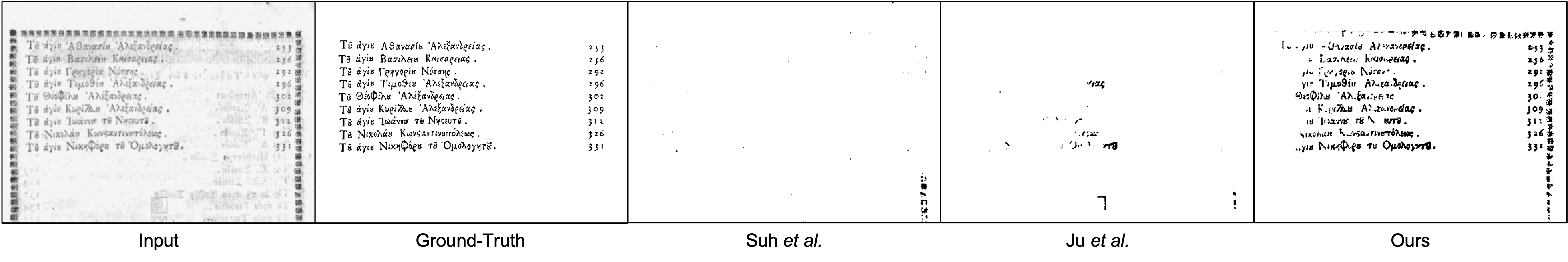}
\vskip 0.01in
\setlength{\tabcolsep}{18pt}{
\begin{tabular}{llcccc}
\toprule
\textbf{Method} & \textbf{Generator} & \textbf{FM$\uparrow$} & \textbf{p-FM$\uparrow$} & \textbf{PSNR$\uparrow$} & \textbf{DRD$\downarrow$} \\ \midrule
Blank Image & -- & -- & -- & 14.19dB & 4.85 \\
Suh \emph{et al.}~\cite{suh2022two} & U-Net \& EfficientNet-B4 & 0.60 & 0.60 & 14.00dB & 5.37 \\
Ju \emph{et al.}~\cite{ju2024three} & U-Net \& EfficientNet-B4 & 10.05 & 9.32 & 14.23dB & 4.98 \\
MFE-GAN (Ours) & U-Net \& EfficientNetV2-S & 56.99 & 56.51 & 14.08dB & 8.38 \\ \bottomrule 
\end{tabular}}
\caption{Representative visualized results from the test set (case 2): from left to right shows input image, ground-truth image, Suh~\emph{et al.}~\cite{suh2022two}, Ju~\emph{et al.}~\cite{ju2024three}, and MFE-GAN (Ours).}
\label{fig:result2}
\end{figure*}

\begin{figure*}[t]
\centering
\includegraphics[width=\linewidth]{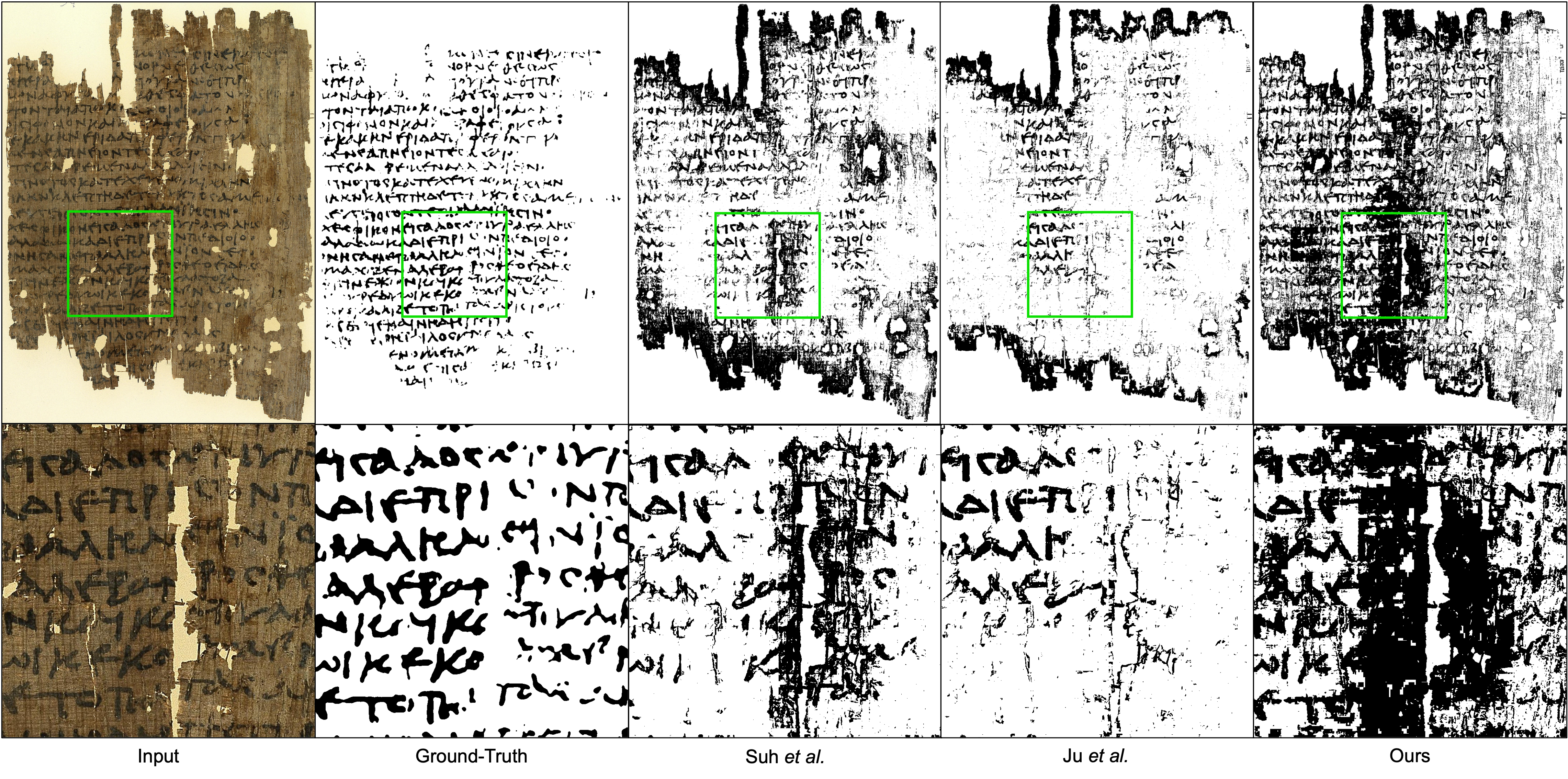}
\vskip 0.01in
\setlength{\tabcolsep}{18pt}{
\begin{tabular}{llcccc}
\toprule
\textbf{Method} & \textbf{Generator} & \textbf{FM$\uparrow$} & \textbf{p-FM$\uparrow$} & \textbf{PSNR$\uparrow$} & \textbf{DRD$\downarrow$} \\ \midrule
Blank Image & -- & -- & -- & 9.71dB & 47.39 \\
Suh \emph{et al.} \cite{suh2022two} & U-Net \& EfficientNet-B4 & 38.46 & 38.68 & 6.75dB & 98.61 \\
Ju \emph{et al.} \cite{ju2024three} & U-Net \& EfficientNet-B4 & 31.58 & 31.66 & 7.90dB & 74.55 \\
MFE-GAN (Ours) & U-Net \& EfficientNetV2-S & 45.49 & 45.58 & 6.14dB & 113.94 \\ \bottomrule 
\end{tabular}}
\caption{Representative visualized results from the test set (case 3): from left to right shows input image, ground-truth image, Suh~\emph{et al.}~\cite{suh2022two}, Ju~\emph{et al.}~\cite{ju2024three}, and MFE-GAN (Ours).}
\label{fig:result3}
\end{figure*}

\subsubsection{Nabuco Dataset}
For the Nabuco dataset, we adopt a two-fold cross-validation strategy, where the final evaluation results are obtained by averaging the outcomes of both validations, as presented in Table~\ref{tab:nabuco}.

MFE-GAN (U-Net++ \& EfficientNetV2-S) achieves the highest FM and p-FM values of 88.04 and 88.72, respectively, as well as the lowest DRD value of 5.06. 
The highest PSNR of 18.65 dB is achieved by the model with U-Net++ \& EfficientNet-B5 of Ju~\emph{et al.}~\cite{ju2024three}, but MFE-GAN with U-Net++ \& EfficientNetV2-S ranks second, achieving 18.60 dB.
Notably, MFE-GAN also obtains the best ASM of 72.58, surpassing the second-best ASM of 72.34 achieved by the model of Ju~\emph{et al.}~\cite{ju2024three} with U-Net++ \& EfficientNet-B5.

Furthermore, our two models rank first and second in terms of the shortest training time for Stage 2 and Stage 3 Top, respectively. 
The total training and inference times for MFE-GAN with U-Net \& EfficientNetV2-S are the shortest, at 23.97h and 0.18h, respectively.
Meanwhile, MFE-GAN with U-Net++ \& EfficientNetV2-S achieves the second shortest total training time of 25.93h.
In conclusion, the comparison outcomes on the Nabuco dataset are consistent with the results observed on the Benchmark Dataset.

\subsubsection{CMATERdb Dataset}
To demonstrate the effectiveness of MFE-GAN in scenarios with limited data, we evaluate it on the CMATERdb dataset, which consists of only five representative images.
We adopt a five-fold cross-validation strategy, indicating that the results in Table~\ref{tab:cmaterdb} are averaged over the five validation runs.
Notably, due to the limited training and test data, we report the training and inference times in seconds (s).

In terms of the FM, p-FM, PSNR, DRD, and ASM evaluation metrics, MFE-GAN with U-Net++ \& EfficientNetV2-S achieves the best performance across all metrics, while Ju \emph{et al.}'s~\cite{ju2024three} model with U-Net++ \& EfficientNet-B5 ranks second.
Furthermore, our top-performing model (U-Net++ \& EfficientNetV2-S) also achieves the second-shortest training and inference times.
In contrast, Ju \emph{et al.}'s~\cite{ju2024three}'s model, despite ranking second in performance, is significantly slower due to its longer Stage 2 training time.
Furthermore, MFE-GAN with U-Net \& EfficientNetV2-S still achieves the shortest training and inference times of 8,894s and 3.06s, respectively, while achieving an ASM value of 73.10.

\subsection{Visual Results\label{sec:visual}}
We randomly select three images from the test set of Benchmark Dataset for visual examination and to demonstrate that the PSNR metric does not directly reflect model performance.

As shown in Figure~\ref{fig:result1}, MFE-GAN generates more complete foreground information.
However, due to the high contamination of the document image, some noise is inevitable when generating additional foreground content.
In contrast, Suh~\emph{et al.}'s~\cite{suh2022two} and Ju~\emph{et al.}'s~\cite{ju2024three}  methods generate less foreground content.
In this case, the PSNR metric fails to intuitively reflect the actual binarization quality.
Specifically, although the FM and p-FM values of the binarized image generated by MFE-GAN are significantly higher than those of the other two methods, its PSNR value (14.96 dB) is very close to that of Suh~\emph{et al.}~\cite{suh2022two} (14.83 dB).

Figure~\ref{fig:result2} further confirms this observation.
Here, a blank image yields a PSNR of 14.19dB, which is higher than that of MFE-GAN (14.08dB).
However, our generated image shows closer visual correspondence to the ground-truth image, indicating that PSNR alone may not fully reflect perceptual quality.

Figure~\ref{fig:result3} presents another case of a lower PSNR value, where MFE-GAN does not process background noise as effectively as the other two methods.
However, MFE-GAN more faithfully generates the original text information compared to others. 
In addition, the PSNR metric is also unreliable in this case because the provided ground-truth itself is flawed, failing to capture the damaged edges of the page.
For instance, a blank image (i.e., all pixels set to white) yields a PSNR of 9.71dB.
This value is significantly higher than MFE-GAN's score of 11.75dB, even though it does not contain any text.

These observations support our claim that a higher PSNR value is not indicative of better model performance, and MFE-GAN can successfully generate more textual information.
Additional qualitative results are provided in Appendix~\ref{appendix:qualitative}.

\begin{table*}[t]
\centering
\caption{Ablation study of each component in the proposed MFE-GAN on the \textbf{Benchmark Dataset}.
The checkmark ($\checkmark$) indicates that the corresponding component is activated. 
The first row corresponds to the full model configuration, as reported in Table~\ref{tab:baseline}.}
\setlength{\tabcolsep}{8pt}{
\begin{tabular}{cccccccc}
\toprule
\multicolumn{5}{c}{\textbf{Component}} & \multicolumn{3}{c}{\textbf{Performance}} \\ \cmidrule(r){1-5} \cmidrule(r){6-8} 
\begin{tabular}[c]{@{}c@{}}\textbf{Backbone}\\EfficientNetV2-S\end{tabular} & \begin{tabular}[c]{@{}c@{}}\textbf{Generator}\\U-Net++\end{tabular} & \begin{tabular}[c]{@{}c@{}}\textbf{Discriminator}\\+ InstanceNorm\end{tabular} & \begin{tabular}[c]{@{}c@{}}\textbf{Generator Loss}\\+ $\lambda_2\mathbb{L}_{\text{Soft-DICE}}$\end{tabular} & \begin{tabular}[c]{@{}c@{}}\textbf{MFE Module}\\HWT\&Norm\end{tabular} & \textbf{ASM} & \textbf{\begin{tabular}[c]{@{}c@{}}Total\\Training\end{tabular}} & \textbf{\begin{tabular}[c]{@{}c@{}}Total\\Inference\end{tabular}} \\ \midrule
\checkmark & \checkmark & \checkmark & \checkmark & \checkmark & \textbf{73.79} & \textbf{68.43h} & \textbf{0.77h} \\ 
 & \checkmark & \checkmark & \checkmark & \checkmark & 73.79 & 112.74h & 1.21h \\
\checkmark &  & \checkmark & \checkmark & \checkmark & 73.23 & 63.91h & 0.68h \\
\checkmark & \checkmark &  & \checkmark & \checkmark & 73.45 & 61.24h & 0.89h \\
\checkmark & \checkmark & \checkmark &  & \checkmark & 73.58 & 70.52h & 0.91h \\
\checkmark & \checkmark & \checkmark & \checkmark &  & 73.81 & 523.86h & 1.19h \\ \bottomrule
\end{tabular}}
\label{tab:ablation}
\end{table*}

\begin{figure*}[t]
\centering
\includegraphics[width=\linewidth]{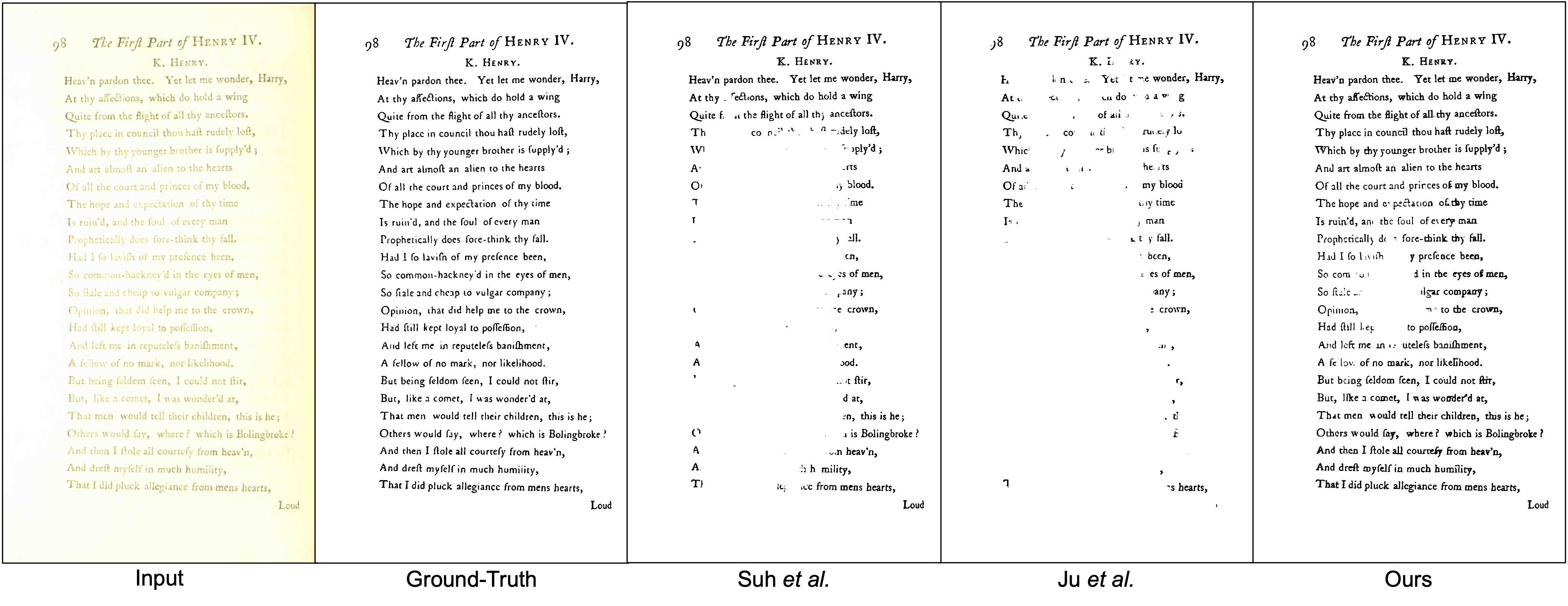}
\caption{Results for a sample affected by overexposure, leading to lower contrast. 
The images from left to right show the input image, ground-truth image, Suh \emph{et al.}~\cite{suh2022two} (U-Net \& EfficientNet-B5), Ju \emph{et al.}~\cite{ju2024three} (U-Net++ \& EfficientNet-B5), and MFE-GAN (U-Net \& EfficientNetV2-S).}
\label{fig:limit}
\end{figure*}

\subsection{Ablation Study\label{sec:ablation}}
To evaluate the contribution of each enhancement in MFE-GAN, we gradually replace or remove each component and observe the impact on performance.
Table~\ref{tab:ablation} summarizes the results under various configurations.
To ensure a fair comparison, all experiments were conducted using the same dataset and hyperparameter settings.

Specifically, replacing U-Net~\cite{ronneberger2015u} with U-Net++~\cite{zhou2019unet++} in the generator and adding instance normalization to the discriminator improve model performance, with a slight increase in training time.
Replacing EfficientNet-B5~\cite{tan2019efficientnet} with EfficientNetV2-S~\cite{tan2021efficientnetv2} in the generator reduces both training and inference times, while the new loss function further improves performance.
Finally, employing the MFE module (i.e., HWT and normalization) for multi-scale feature extraction significantly reduces training time from 523.86h to 68.43h, representing an 87\% decrease, with only a marginal decrease of 0.02 in ASM.

Overall, each improvement objectively contributes to either performance gains or reductions in training and inference times.

\section{Discussion}
\label{sec:discussion}
We select a low-contrast sample with over-exposure from the Benchmark Dataset to showcase the visual results of different methods, and to discuss the limitations of MFE-GAN. 
As shown in Figure~\ref{fig:limit}, the input image is a yellowish manuscript, with the light source on the left side being excessively bright during capture.
This results in a very low contrast between the text and background, making effective image binarization challenging.

Compared to Suh~\emph{et al.}'s~\cite{suh2022two} and Ju~\emph{et al.}'s~\cite{ju2024three} methods, MFE-GAN generates more detailed text.
However, our final results remain unsatisfactory, with text still missing in the central region.
We consider that, when the contrast between the text and background is too low, GANs trained independently on different color channels struggle to effectively distinguish the text from the background, which in turn affects the generation results.

Therefore, for documents affected by light pollution, we suggest that applying contrast enhancement techniques based on both global and local features could help GANs distinguish the text from the background more effectively.
Furthermore, incorporating pre-processing methods based on exposure compensation may mitigate the impact of this issue.
Moreover, we suspect that MFE-GAN may have limited generalization performance under extreme conditions, such as over-exposure and low-contrast scenes, as the Benchmark Dataset contains few light-polluted samples.
To address this, we suggest employing a data augmentation strategy to increase the proportion of such samples under extreme conditions in the training set and thereby enhance the model's generalization ability.

\section{Conclusion and Future Work}
\label{sec:conclusion}
Degraded color document image enhancement and binarization are important steps in document analysis.
Current SOTA GAN-based methods can generate satisfactory document binarization results but suffer from long training and inference times.
To address this drawback, we propose MFE-GAN, an efficient three-stage GAN-based framework that incorporates an MFE module (i.e., HWT and normalization) for multi-scale feature extraction, which significantly reduces training and inference times.
Furthermore, we introduce novel generators, discriminators, and a new loss function to further improve the performance of our proposed MFE-GAN.
Experimental results on benchmark datasets demonstrate that MFE-GAN not only achieves superior model performance but also significantly reduces the total training and inference times in comparison to SOTA GAN-based methods.

For future work, we explore the integration of document image binarization and document image understanding for practical applications, especially for ancient documents or historical artifacts.
Such applications could include real-time translation, summarization, and related document retrieval.

\printcredits
\section*{Declarations}
This paper is an expanded version of the paper~\cite{ju2025efficient} presented at the 17th Asia Pacific Signal and Information Processing Association Annual Summit and Conference (APSIPA ASC), held from October 22 to 24, 2025, in Singapore.

\section*{Acknowledgment}
This research is supported by National Science and Technology Council of Taiwan, under Grant Number: NSTC 114-2221-E-032-011-.

\section*{Declaration of competing interest}
The authors have no financial or proprietary interests in any material discussed in this article.

\section*{Declaration of Generative AI and AI-assisted technologies in the writing process}
The authors only use generative artificial intelligence (AI) and AI-assisted technologies to improve language.

\bibliographystyle{elsarticle-num}
\bibliography{cas-refs}


\clearpage
\appendix

\begin{table*}[t]
\centering
\caption{Quantitative comparison of MFE-GAN using different architectures and backbone models to construct the generator.}
\setlength{\tabcolsep}{7.5pt}{
\begin{tabular}{lcccccccc}
\toprule
\textbf{Generator} & \textbf{FM$\uparrow$} & \textbf{p-FM$\uparrow$} & \textbf{PSNR$\uparrow$} & \textbf{DRD$\downarrow$} & \textbf{ASM$\uparrow$} & \textbf{Total Training$\downarrow$} & \textbf{Total Inference$\downarrow$} \\ \cmidrule(r){1-1} \cmidrule(r){2-5} \cmidrule(r){6-8}
U-Net \& EfficientNetV2-S & 88.83 & \textbf{89.87} & \textbf{19.07dB} & 4.86 & \textbf{73.23} & \textbf{63.91h} & \textbf{0.68h} \\
U-Net \& EfficientNet-B4 & 87.87 & 88.57 & 18.82dB & 5.17 & 72.52 & 69.38h & 0.75h \\ 
U-Net \& EfficientNet-B5 & \textbf{88.88} & 89.65 & 18.93dB & \textbf{4.85} & 73.15 & 81.63h & 0.83h \\ \cmidrule(r){1-1} \cmidrule(r){2-5} \cmidrule(r){6-8}
U-Net++ \& EfficientNetV2-S & 89.69 & \textbf{90.78} & \textbf{19.15dB} & \textbf{4.45} & \textbf{73.79} & \textbf{68.43h} & \textbf{0.77h} \\
U-Net++ \& EfficientNet-B4 & 89.40 & 90.38 & 19.01dB & 4.87 & 73.48 & 85.45h & 0.91h \\ 
U-Net++ \& EfficientNet-B5 & \textbf{89.76} & 90.75 & \textbf{19.15dB} & 4.51 & \textbf{73.79} & 112.74h & 1.21h \\ \bottomrule
\end{tabular}}
\begin{minipage}{0.95\textwidth}
\footnotesize
The best value in each group is highlighted in \textbf{bold}.
\end{minipage}
\label{tab:ablation_arch}
\end{table*}

\begin{table*}[t]
\centering
\caption{Training and inference times of the original models before (Baseline) and after applying the MFE module (Ours).}
\setlength{\tabcolsep}{9.5pt}{
\begin{tabular}{lcccccc}
\toprule
\textbf{Method} & 
\textbf{\begin{tabular}[c]{@{}c@{}}Stage~2\\ Train\end{tabular}} & 
\textbf{\begin{tabular}[c]{@{}c@{}}Stage~2\\ Predict\end{tabular}} & 
\textbf{\begin{tabular}[c]{@{}c@{}}Stage~3\\ Top\end{tabular}} & 
\textbf{\begin{tabular}[c]{@{}c@{}}Stage~3\\ Bottom\end{tabular}} & 
\textbf{\begin{tabular}[c]{@{}c@{}}Total\\ Training\end{tabular}} & 
\textbf{\begin{tabular}[c]{@{}c@{}}Total\\ Inference\end{tabular}} \\ 
\cmidrule(r){1-1} \cmidrule(r){2-5} \cmidrule(r){6-7}
U-Net \& EfficientNetV2-S (Baseline) & 332.28h & 3.56h & \textbf{47.47h} & 1.63h & 384.95h & 1.12h \\ 
U-Net \& EfficientNetV2-S (MFE-GAN) & \textbf{11.60h} & \textbf{3.45h} & \textbf{47.47h} & \textbf{1.39h} & \textbf{63.91h} & \textbf{0.68h} \\ 
\cmidrule(r){1-1} \cmidrule(r){2-5} \cmidrule(r){6-7}
U-Net++ \& EfficientNetV2-S (Baseline) & 465.28h & 3.94h & 52.88h & 1.76h & 523.86h & 1.19h \\
U-Net++ \& EfficientNetV2-S (MFE-GAN) & 14.12h & 3.63h & 49.29h & \textbf{1.39h} & 68.43h & 0.77h \\ 
\bottomrule
\end{tabular}}
\begin{minipage}{0.95\textwidth}
\footnotesize
The best value in each group is highlighted in \textbf{bold}.
\end{minipage}
\label{tab:ablation_time}
\end{table*}

\begin{table*}[t]
\centering
\caption{Quantitative comparison of our method with different loss function configurations.}
\setlength{\tabcolsep}{7.5pt}{
\begin{tabular}{cccccccc}
\toprule
\textbf{Generator Loss Function} & \textbf{FM$\uparrow$} & \textbf{p-FM$\uparrow$} & \textbf{PSNR$\uparrow$} & \textbf{DRD$\downarrow$} & \textbf{ASM$\uparrow$} & \textbf{Total Train$\downarrow$} & \textbf{Total Inference$\downarrow$}\\ \cmidrule(r){1-1} \cmidrule(r){2-5} \cmidrule(r){6-8}
$D(G(z))$ + $\lambda\mathbb{L}_{\text{BCE}}$ & 89.41 & 90.42 & 19.10dB & 4.61 & 73.58 & 70.52h & 0.91h \\
$D(G(z))$ + $\lambda\mathbb{L}_{\text{Soft-DICE}}$ & 88.81 & 90.00 & 19.09dB & \textbf{4.31} & 73.40 & 71.55h & \textbf{0.75h} \\
$D(G(z))$ + $\lambda_1\mathbb{L}_{\text{BCE}}$ + $\lambda_2\mathbb{L}_{\text{Soft-DICE}}$ & \textbf{89.69} & \textbf{90.78} & \textbf{19.15dB} & 4.45 & \textbf{73.79} & \textbf{68.43h} & 0.77h \\ \bottomrule
\end{tabular}}
\begin{minipage}{0.95\textwidth}
\footnotesize
We set $\lambda = 50$ and $\lambda_1 = \lambda_2 = 25$.
The best value in each group is highlighted in \textbf{bold}.
\end{minipage}
\label{tab:loss}
\end{table*}

\begin{table*}[t]
\centering
\caption{Quantitative comparison (ASM: FM/p-FM/PSNR/DRD) of the proposed MFE-GAN and other methods on each DIBCO dataset.}
\subfigure[DIBCO 2011]{
\setlength{\tabcolsep}{5.5pt}{
\begin{tabular}{lcccc}
\toprule
\textbf{Method} & \textbf{FM$\uparrow$} & \textbf{p-FM$\uparrow$} & \textbf{PSNR$\uparrow$} & \textbf{DRD$\downarrow$} \\ \midrule
Otsu~\cite{otsu1979threshold} & 82.10& 85.96& 15.72dB&8.95\\
Sauvola~\cite{sauvola2000adaptive} & 82.14& 87.70& 15.65dB&8.50\\ 
U-Net \& B4~\cite{suh2022two} & 89.38 & 90.44 & 19.71dB & 3.25 \\
U-Net \& B5~\cite{suh2022two} & 89.64 & 91.24 & 19.76dB & 3.02 \\
U-Net++ \& B4~\cite{ju2024three} & 89.02 & 89.96 & 19.67dB & 3.02 \\
U-Net++ \& B5~\cite{ju2024three} & 91.89 & {\color{red}93.58} & 19.73dB & 2.95 \\
\textbf{U-Net \& V2-S} & {\color{blue}92.47} & 93.14 & {\color{blue}19.77dB} & {\color{blue}2.81} \\
\textbf{U-Net++ \& V2-S} & {\color{red}92.83} & {\color{blue}93.50} & {\color{red}19.92dB} & {\color{red}2.58} \\ \bottomrule
\end{tabular}}}
\subfigure[DIBCO 2013]{
\setlength{\tabcolsep}{5.5pt}{
\begin{tabular}{lcccc}
\toprule
\textbf{Method} & \textbf{FM$\uparrow$} & \textbf{p-FM$\uparrow$} & \textbf{PSNR$\uparrow$} & \textbf{DRD$\downarrow$} \\ \midrule
Otsu~\cite{otsu1979threshold} & 80.04& 83.43& 16.63dB&10.98\\
Sauvola~\cite{sauvola2000adaptive} & 82.71& 87.74& 17.02dB&7.64\\ 
U-Net \& B4~\cite{suh2022two} & 93.23 & 93.30 & 20.81dB & 2.88 \\
U-Net \& B5~\cite{suh2022two} & {\color{blue}94.23} & {\color{blue}94.68} & {\color{red}21.48dB} & {\color{blue}2.46} \\ 
U-Net++ \& B4~\cite{ju2024three} & 93.81 & 94.19 & 21.06dB & 2.68 \\ 
U-Net++ \& B5~\cite{ju2024three} & {\color{red}94.34} & {\color{red}94.72} & {\color{blue}21.28dB} & {\color{red}2.17} \\ 
\textbf{U-Net \& V2-S} & 92.80 & 93.18 & 20.98dB & 3.19 \\ 
\textbf{U-Net++ \& V2-S} & 93.23 & 93.57 & 21.09dB & 2.77 \\ \bottomrule
\end{tabular}}}
\subfigure[H-DIBCO 2014]{
\setlength{\tabcolsep}{5.5pt}{
\begin{tabular}{lcccc}
\toprule
\textbf{Method} & \textbf{FM$\uparrow$} & \textbf{p-FM$\uparrow$} & \textbf{PSNR$\uparrow$} & \textbf{DRD$\downarrow$} \\ \midrule
Otsu~\cite{otsu1979threshold} & 91.62& 95.69& 18.72dB&2.65\\
Sauvola~\cite{sauvola2000adaptive} & 84.70& 87.88& 17.81dB&4.77\\ 
U-Net \& B4~\cite{suh2022two} & 96.19 & 96.71 & 21.58dB & 1.15 \\
U-Net \& B5~\cite{suh2022two} & {\color{blue}96.37} & 96.90 & 21.78dB & {\color{blue}1.09} \\
U-Net++ \& B4~\cite{ju2024three} & 95.96 & 96.33 & 21.31dB & 1.22 \\
U-Net++ \& B5~\cite{ju2024three} & {\color{red}96.38} & {\color{blue}96.96} & {\color{blue}21.85dB} & {\color{red}1.08} \\ 
\textbf{U-Net \& V2-S} & 96.28 & 96.77 & 21.78dB & 1.13 \\ 
\textbf{U-Net++ \& V2-S} & 96.36 & {\color{red}97.72} & {\color{red}21.91dB} & {\color{red}1.08} \\ \bottomrule
\end{tabular}}}
\subfigure[H-DIBCO 2016]{
\setlength{\tabcolsep}{5.5pt}{
\begin{tabular}{lcccc}
\toprule
\textbf{Method} & \textbf{FM$\uparrow$} & \textbf{p-FM$\uparrow$} & \textbf{PSNR$\uparrow$} & \textbf{DRD$\downarrow$} \\ \midrule
Otsu~\cite{otsu1979threshold} & 86.59& 89.92& 17.79dB&5.58\\
Sauvola~\cite{sauvola2000adaptive} & 84.64& 88.39& 17.09dB&6.27\\ 
U-Net \& B4~\cite{suh2022two} & 91.91 & 95.00 & 19.67dB & 2.99 \\ 
U-Net \& B5~\cite{suh2022two} & 91.97 & {\color{red}95.23} & 19.69dB & 2.94 \\
U-Net++ \& B4~\cite{ju2024three} & {\color{blue}92.31} & 94.86 & {\color{blue}19.83dB} & {\color{blue}2.80} \\ 
U-Net++ \& B5~\cite{ju2024three} & {\color{red}92.42} & 95.03 & {\color{red}19.87dB} & {\color{red}2.79} \\ 
\textbf{U-Net \& V2-S} & 91.82 & 94.17 & 19.53dB & 2.97 \\ 
\textbf{U-Net++ \& V2-S} & 92.05 & 94.26 & 19.62dB & 2.85 \\ \bottomrule
\end{tabular}}}
\subfigure[DIBCO 2017]{
\setlength{\tabcolsep}{5.5pt}{
\begin{tabular}{lcccc}
\toprule
\textbf{Method} & \textbf{FM$\uparrow$} & \textbf{p-FM$\uparrow$} & \textbf{PSNR$\uparrow$} & \textbf{DRD$\downarrow$} \\ \midrule
Otsu~\cite{otsu1979threshold} & 77.73& 77.89& 13.85dB&15.54\\
Sauvola~\cite{sauvola2000adaptive} & 77.11& 84.10& 14.25dB&8.85\\ 
U-Net \& B4~\cite{suh2022two} & {\color{blue}89.65} & {\color{blue}90.42} & 17.76dB & 3.82 \\
U-Net \& B5~\cite{suh2022two} & {\color{red}89.89} & {\color{red}90.97} & {\color{blue}17.95dB} & {\color{red}3.61} \\
U-Net++ \& B4~\cite{ju2024three} & 87.58 & 88.39 & 17.58dB & 4.67 \\
U-Net++ \& B5~\cite{ju2024three} & 88.72 & 89.81 & {\color{red}17.97dB} & {\color{blue}3.77} \\
\textbf{U-Net \& V2-S} & 88.43 & 89.36 & 17.60dB & 4.04 \\
\textbf{U-Net++ \& V2-S} & 89.34 & 90.14 & 17.64dB & 3.80 \\ \bottomrule
\end{tabular}}}
\subfigure[H-DIBCO 2018]{
\setlength{\tabcolsep}{5.5pt}{
\begin{tabular}{lcccc}
\toprule
\textbf{Method} & \textbf{FM$\uparrow$} & \textbf{p-FM$\uparrow$} & \textbf{PSNR$\uparrow$} & \textbf{DRD$\downarrow$} \\ \midrule
Otsu~\cite{otsu1979threshold} & 51.45& 53.05& 9.74dB&59.07\\
Sauvola~\cite{sauvola2000adaptive} & 67.81& 74.08& 13.78dB&17.69\\ 
U-Net \& B4~\cite{suh2022two} & 93.53 & 95.17 & 20.59dB & 2.23 \\
U-Net \& B5~\cite{suh2022two} & {\color{blue}93.69} & {\color{red}95.58} & {\color{red}20.74dB} & {\color{blue}2.16} \\ 
U-Net++ \& B4~\cite{ju2024three} & 93.58 & 94.73 & 20.50dB & 2.20 \\ 
U-Net++ \& B5~\cite{ju2024three} & {\color{red}93.75} & {\color{blue}95.54} & {\color{blue}20.71dB} & {\color{red}2.11} \\ 
\textbf{U-Net \& V2-S} & 92.85 & 94.65 & 20.14dB & 2.71 \\ 
\textbf{U-Net++ \& V2-S} & 93.61 & 95.28 & 20.07dB & 2.60 \\ \bottomrule
\end{tabular}}}
\subfigure[DIBCO 2019]{
\setlength{\tabcolsep}{5.5pt}{
\begin{tabular}{lcccc}
\toprule
\textbf{Method} & \textbf{FM$\uparrow$} & \textbf{p-FM$\uparrow$} & \textbf{PSNR$\uparrow$} & \textbf{DRD$\downarrow$} \\ \midrule
Otsu~\cite{otsu1979threshold} & 47.83& 45.59& 9.08dB&109.46\\
Sauvola~\cite{sauvola2000adaptive} & 51.73& 55.15& 13.72dB&13.83\\ 
U-Net \& B4~\cite{suh2022two} & 61.76 & 62.00 & 13.58dB & 17.46 \\
U-Net \& B5~\cite{suh2022two} & 64.11 & 64.74 & {\color{blue}13.77dB} & 15.97 \\
U-Net++ \& B4~\cite{ju2024three} & 64.75 & {\color{blue}69.49} & 13.65dB & {\color{blue}15.87} \\
U-Net++ \& B5~\cite{ju2024three} & 66.42 & 66.80 & 13.70dB & 16.53 \\
\textbf{U-Net \& V2-S} & {\color{blue}67.17} & 67.83 & 13.70dB & 17.17 \\
\textbf{U-Net++ \& V2-S} & {\color{red}70.41} & {\color{red}70.96} & {\color{red}13.79dB} & {\color{red}15.49} \\ \bottomrule
\end{tabular}}}
\subfigure[Mean Values]{
\setlength{\tabcolsep}{5.5pt}{
\begin{tabular}{lcccc}
\toprule
\textbf{Method} & \textbf{FM$\uparrow$} & \textbf{p-FM$\uparrow$} & \textbf{PSNR$\uparrow$} & \textbf{DRD$\downarrow$} \\ \midrule
Otsu~\cite{otsu1979threshold} & 73.91& 75.93& 14.50dB&30.32\\
Sauvola~\cite{sauvola2000adaptive} & 75.83& 80.72& 15.62dB&9.65\\ 
U-Net \& B4~\cite{suh2022two} & 87.95 & 89.01 & 19.10dB & 4.83 \\ 
U-Net \& B5~\cite{suh2022two} & 88.56 & 89.90 & {\color{red}19.31dB} & {\color{red}4.46} \\
U-Net++ \& B4~\cite{ju2024three} & 88.14 & 89.71 & 19.09dB & 4.64 \\ 
U-Net++ \& B5~\cite{ju2024three} & {\color{blue}89.13} & {\color{blue}90.35} & {\color{blue}19.30dB} & 4.49 \\ 
\textbf{U-Net \& V2-S} & 88.83 & 89.87 & 19.07dB & 4.86 \\ 
\textbf{U-Net++ \& V2-S} & {\color{red}89.69} & {\color{red}90.78} & 19.15dB & {\color{blue}4.45} \\ \bottomrule
\end{tabular}}}
\begin{minipage}{0.95\textwidth}
\footnotesize
The proposed MFE-GAN is highlighted in \textbf{bold}.
The best and second-best performances are highlighted in {\color{red}red} and {\color{blue}blue}, respectively.
Since papers~\cite{suh2022two,ju2024three} did not provide experimental results on all datasets for the configuration shown in the table, we have independently trained these models ourselves.
\end{minipage}
\label{tab:result}
\end{table*}

\begin{figure*}[t]
\centering
\includegraphics[width=\linewidth]{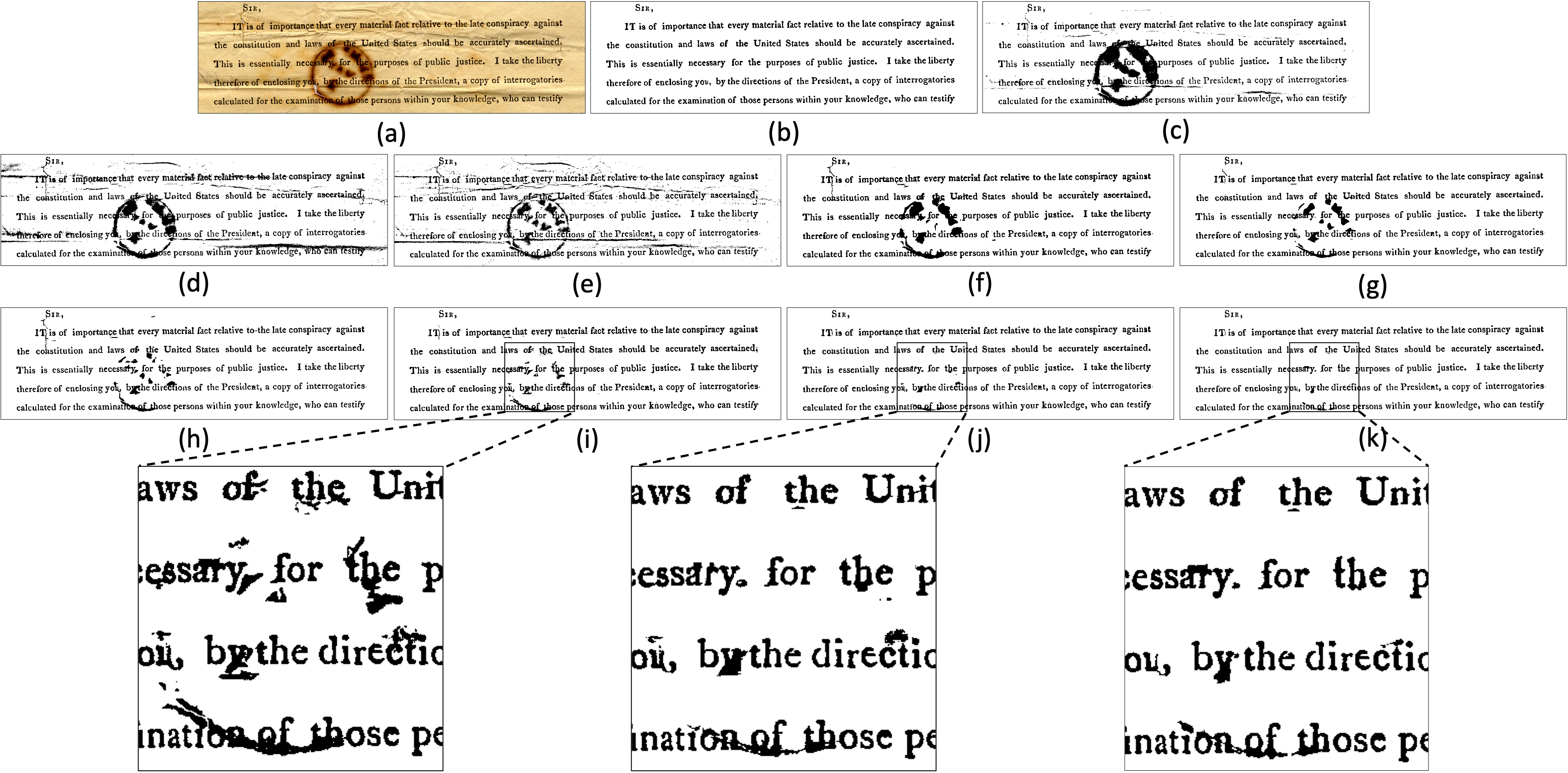}
\caption{Visual comparison of binarization methods on a sample from the DIBCO 2013 dataset: 
(a) Input, (b) Ground-Truth, (c) Otsu~\cite{otsu1979threshold}, (d) Niblack~\cite{niblack1985introduction}, (e) Sauvola~\cite{sauvola2000adaptive}, (f) Vo~\cite{vo2018binarization}, (g) He~\cite{he2019deepotsu}, (h) Zhao~\cite{zhao2019document}, (i) Suh~\cite{suh2021cegan}, (j) Ju~\cite{ju2024three}, (k) MFE-GAN.}
\label{fig:visual1}
\end{figure*}

\begin{figure*}[t]
\centering
\includegraphics[width=\linewidth]{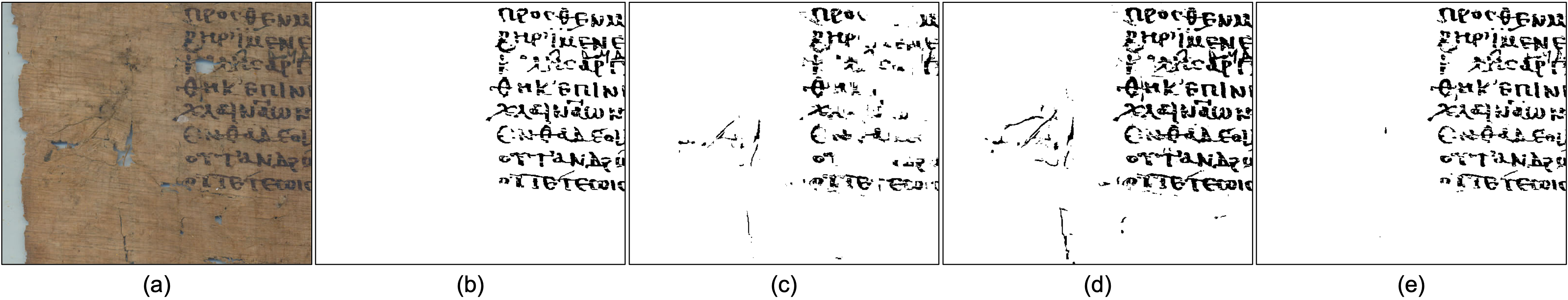}
\caption{Visual comparison of binarization methods on a sample from the DIBCO 2019 dataset: (a) Input Image, (b) Ground-Truth, (c) Suh~\cite{suh2022two}, (d) Ju~\cite{ju2024three}, (e) MFE-GAN.}
\label{fig:visual2}
\end{figure*}

\section{Effect on the Generator Architecture\label{appendix:generator}}
One significant contribution of this work is the design of novel generators for the proposed three-stage GAN framework.
This work aims to enable the trained model to generate more foreground text information using these novel generators. 
To demonstrate that our proposed generator backbone (EfficientNetV2~\cite{tan2021efficientnetv2}) is superior to the original backbone (EfficientNet~\cite{tan2019efficientnet}) when it is used together with U-Net~\cite{ronneberger2015u} or U-Net++~\cite{zhou2019unet++}, we conduct a series of experiments. 

As shown in Table~\ref{tab:ablation_arch}, we evaluate the model performance as well as the total training and inference times across different generator architectures.
For the encoder of the generators, we utilize EfficientNet-B4, EfficientNet-B5, and EfficientNetV2-S.
Table~\ref{tab:ablation_arch} indicates that, for different GAN encoders, the proposed MFE-GAN achieves shorter total training and inference times than the original models, while maintaining comparable or higher ASM values.

Specifically, the original method using U-Net++~\cite{zhou2019unet++} with EfficientNet-B5~\cite{tan2019efficientnet} achieves an ASM value of 73.79, with training and inference times of 112.74h and 1.21h, respectively.
In contrast, MFE-GAN achieves the same ASM value but with a total training time of 68.43h and a total inference time of 0.77h, which represents a decrease of 39\% and 36\%, respectively. 
These experiment results demonstrate that the proposed MFE-GAN can greatly reduce the training and inference times while maintaining, or even improving, model performance.

\section{Effect on the MFE Module}\label{appendix:mfe}
To demonstrate the effectiveness of the MFE module (i.e., applying HWT and normalization) in Stage~1, we evaluate two configurations of MFE-GAN:
Model~A: U-Net~\cite{ronneberger2015u} with EfficientNetV2-S~\cite{tan2021efficientnetv2}, and Model~B: U-Net++~\cite{zhou2019unet++} with EfficientNetV2-S~\cite{tan2021efficientnetv2}. 

Table~\ref{tab:ablation_time} summarizes the time taken for each stage, as well as the total training and inference times of the compared methods.
Two configurations are compared: one with the application of the MFE module in Stage~1 (processing 128 $\times$ 128 sub-bands), and one without (i.e., the \emph{baseline} method, where the original 256 $\times$ 256 patches are directly fed into the GANs).

Here, the total training time refers to the sum of the durations of all stages, while the total inference time denotes the time required to generate images for all test sets.
It can be seen that for both models, the total training time decreases when the MFE module is applied.

Specifically, the training time decreases from 384.95h to 63.91h for Model~A and from 523.86h to 68.43h for Model~B when the MFE module is applied.
Similarly, the total inference time decreases from 1.12h to 0.68h for Model~A and from 1.19h to 0.77h for Model~B.
These results demonstrate that incorporating the MFE module significantly reduces both training and inference times.

\section{Effect on the Loss Functions}\label{appendix:loss}
To validate that combining both BCE loss and Soft Dice loss in the generator’s loss function can improve model performance, we conduct a comparative experiment.
Specifically, we adopt the loss function $D(G(z))+0.5\times \mathop{\mathbb{L}_{\text{BCE}}}$, as used in~\cite{suh2022two,ju2024three}, as our baseline (see Table~\ref{tab:baseline}). 
In addition, we introduce two additional configurations: (1) replacing BCE loss with Soft Dice loss: $D(G(z))+0.5\times \mathop{\mathbb{L}_{\text{Soft-DICE}}}$; and (2) combining BCE loss and Soft Dice loss:  $D(G(z))+0.25\times \mathop{\mathbb{L}_{\text{BCE}}}+0.25\times \mathop{\mathbb{L}_{\text{Soft-DICE}}}$.

As presented in Table~\ref{tab:loss}, the experimental results show that directly replacing BCE loss with Soft Dice loss leads to a decline in model performance.
This demonstrates that BCE loss contributes positively to model performance in binary classification (text vs. background).
In contrast, MFE-GAN, which combines both BCE loss and Soft Dice loss, achieves the best performance across FM, p-FM, and PSNR evaluation metrics.
In addition, it outperforms the baseline model, achieving the shortest total training time and a reduced total inference time.

\section{Results on Each DIBCO Dataset}\label{appendix:dibco}
Table~\ref{tab:comparison_performance} reports the average values of each metric for different models on the Benchmark Dataset.
Since the Benchmark Dataset consists of several DIBCO datasets, the results for each DIBCO dataset are shown in Table~\ref{tab:result}.
MFE-GAN (using U-Net++ \& EfficientNetV2-S) achieves the best or second-best results on several datasets, including DIBCO~2011, H-DIBCO~2014, and DIBCO~2019, demonstrating its robustness to various types of document degradation.
Overall, MFE-GAN achieves the highest mean FM and p-FM values (89.69\% and 90.78\%), along with a competitive DRD value of 4.45, outperforming the previous model (U-Net++ \& EfficientNet-B5) proposed in~\cite{ju2024three}.

\section{Qualitative Comparison}\label{appendix:qualitative}
Beyond the quantitative comparisons on different datasets, Figures~\ref{fig:visual1} and~\ref{fig:visual2} compares the binarized images produced by using different methods for a sample from the DIBCO~2013 and 2019 datasets, respectively. 
These figures demonstrate that SOTA GAN-based methods outperform traditional binarization methods in terms of shadow and noise elimination.
Furthermore, MFE-GAN excels in preserving textual content while effectively mitigating shadows and noise.

\end{document}